\documentclass[sigconf]{acmart}

\usepackage{graphicx}
\usepackage{float}
\usepackage{amsmath,mathtools}

\usepackage{amssymb}
\usepackage{bm}
\usepackage{booktabs}
\usepackage{siunitx}
\usepackage{capt-of,placeins}
\usepackage{enumitem}
\usepackage{subcaption}
\usepackage{multirow}
\usepackage{caption}
\usepackage[ruled,vlined]{algorithm2e}

\usepackage{tikz}
\usetikzlibrary{arrows.meta, positioning, fit}
\usepackage{dblfloatfix}

\setcopyright{rightsretained}
\acmYear{2025}
\acmConference[KDD Workshop ’25]
  {KDD Workshop on Prompt Optimization}
  {August 4, 2025}
  {Toronto, ON, Canada}
\acmBooktitle{Proceedings of the KDD Workshop on Prompt Optimization}

\AtBeginDocument{%
  \providecommand\BibTeX{{\normalfont B\kern-0.5em{\scshape i\kern-0.25em b}\kern-0.8em\TeX}}%
}

\title{DeRAG: Black-box Adversarial Attacks on Multiple Retrieval-Augmented Generation Applications via Prompt Injection }

\author{Jerry Wang}
\affiliation{%
  \institution{Department of Management Information Systems, National ChengChi University}
  \city{Taipei}
  \country{Taiwan}
}
\email{111306078@nccu.edu.tw}

\author{Fang Yu}
\affiliation{%
  \institution{Department of Management Information Systems, National ChengChi University}
  \city{Taipei}
  \country{Taiwan}
}
\email{yuf@nccu.edu.tw}

\begin{document}  

\begin{abstract}
Adversarial prompt attacks can significantly alter the reliability of Retrieval-Augmented Generation (RAG) systems by re-ranking them to produce incorrect outputs. In this paper, we present a novel method that applies Differential Evolution (DE) to optimize adversarial prompt suffixes for RAG-based question answering. Our approach is gradient-free, treating the RAG pipeline as a black box and evolving a population of candidate suffixes to maximize the retrieval rank of a targeted incorrect document to be  closer to real world scenarios. 
We conducted experiments on the BEIR QA datasets to evaluate attack success at certain retrieval rank thresholds under multiple retrieving applications. Our results demonstrate that DE-based prompt optimization attains competitive (and in some cases higher) success rates compared to GGPP to dense retrievers and PRADA to sparse retrievers, while using only a small number of tokens $(\leq 5\ \text{tokens})$ in the adversarial suffix. Furthermore, we introduce a readability‐aware suffix construction strategy, validated by a statistically significant reduction in MLM negative log‐likelihood with Welch’s $t$-test.
Through evaluations with a BERT‐based adversarial suffix detector, we show that DE‐generated suffixes evade detection, yielding near‐chance detection accuracy. 
\end{abstract}

\ccsdesc[500]{Information systems~Retrieval models and ranking}

\keywords{Prompt injection, Differential evolution, RAG attack}

\maketitle

\section{Introduction}
Retrieval-Augmented Generation (RAG) combines large language models (LLMs) with information retrieval to ground LLM outputs in external documents~\cite{Hu2024GGPP,lewis2021nlp,li2024enhancingllmfactualaccuracy}. 
By retrieving relevant passages from a corpus to include as context, RAG aims to improve the factual accuracy and reduce hallucinations in generated answers
. 
Recent advances in RAG systems have placed increasing emphasis on the quality of retrieval, particularly the use of powerful embedding models. State-of-the-art retrievers now leverage instruction-tuned embeddings~\cite{muennighoff2024grit} or compact, high-performing open-source  models~\cite{merrick2024arcticembed}, which significantly improve retrieval precision across diverse tasks. Jha et al.\ ~\cite{jha2025harnessing} further demonstrate that embeddings from diverse models 
(e.g., BERT, RoBERTa, CLIP) reside in a common latent semantic geometry. Their unsupervised vec2vec mapping provided strong support for a universal latent space across embedding architectures. Moreover, emerging unified models such as GritLM~\cite{muennighoff2024grit} enable a single large language model to perform both retrieval and generation efficiently, reducing inference latency and simplifying the deployment pipeline. 

However, recent work has revealed that RAG pipelines remain vulnerable to adversarial input manipulations. Li \emph{et al.} ~\cite{li2024} show that even a deceptively simple adversarial prefix can subvert RAG-based AI agents by bypassing LLM safeguards and forcing dangerous or unintended outputs. Xue \emph{et al.}~\cite{xue2024} propose \textsc{TrojRAG}, a poisoning‐based backdoor on RAG databases that, with only a handful of crafted passages, can reliably hijack retrieval and severely undermine downstream LLM performance. In particular, inserting a carefully crafted prefix or suffix into a user’s query can dramatically alter the retrieved documents, causing the LLM to produce an incorrect specific answer~\cite{sui2025ctrlrag}. This sabotages the trustworthiness of RAG, as malicious actors could exploit such prompts to inject misinformation. Hu et al.~\cite{Hu2024GGPP} introduced Gradient-Guided Prompt Perturbation (GGPP), a white-box attack on RAG pipelines that optimizes a small continuous prefix in the LLM retriever’s embedding space via gradient descent to push a targeted wrong passage into the top retrieval results. Specifically, GGPP minimizes the distance between the query embedding and the target passage embedding while maximizing the distance to the original relevant passage, uses a heuristic prefix initialization from important target‐passage tokens, and projects the optimized embedding back to discrete tokens, achieving high success rates—often ranking the incorrect passage at position 1—and even enabling analysis of internal activations and attack detection. 
Despite GGPP’s success, its gradient‑based nature requires access to the differentiable components of the RAG model (for example, the retriever or encoder gradients). In many real‑world scenarios, the internal model may be a black box (e.g., a closed-source API), or gradient access is not available. This motivates exploring gradient‑free adversarial prompt optimization. 

In this paper, we introduce a Differential Evolution (DE) ~\cite{DEReview2021} based method to generate adversarial prompt suffixes for RAG systems, where we treat the retriever as a black box, requires no gradient access or model internals. Differential Evolution (DE) is a population-based evolutionary optimization algorithm known for efficient black‑box optimization and global search capabilities which is widely used in different scenarios. 
We propose Differential Evolution Prompt Optimization (DeRAG), treating each candidate suffix of length $L$ as an individual in a DE population and evolving them through mutation, crossover, and selection. At each generation, DeRAG evaluates the fitness of a suffix by measuring how effectively it re-ranks a target document to the top of the retrieval list, using only forward calls to the encoder and retriever (i.e., cosine similarities over CLS embeddings). This black-box fitness evaluation makes DeRAG applicable even when model internals are inaccessible.
We evaluate DeRAG on the BEIR benchmarks~\cite{Thakur2021Beir}, specify in MS MARCO~\cite{bajaj2018msmarcohumangenerated} SciFact ~\cite{SciFact2020}, FiQA ~\cite{Maia2018Fiqa} and Fever ~\cite{thorne2018fever}, using a BERT-base-uncased retriever~\cite{devlin2018bert} to extract embeddings. With a budget of only a few hundred model calls, DeRAG achieves high success rates at Top-1, Top-10, and Top-20 retrieval thresholds—matching or surpassing GGPP and outperforming both random and other baselines. We further analyze how the adversarial  suffix length affects attack success and retrieval disruption. We find that short suffixes are in fact sufficient to achieve a high success rate  while larger suffixes yield diminishing marginal returns under same max iterations. We also provide insights into how different DE variants balance efficiency  with success, and we qualitatively observe that the optimized adversarial prompts often consist of obscure or foreign tokens that are semantically unrelated to the query – a strategy that exploits the retriever’s embedding space to confuse retrieval. On top of that, we also discussed the positional differences, which shows the potential of positioning attack. These results demonstrate that evolutionary, gradient-free optimization poses a practical and potent threat to RAG deployments and applications.

\section{Related Work}
\subsection{Adversarial Prompts for LLMs and RAG}
Adversarial examples have long posed a threat to deep neural networks in vision and language domains. In the context of large language models (LLMs), adversarial prompts represent a textual counterpart—crafted inputs designed to manipulate model behavior without altering the underlying parameters. Adversarial attacks on large language models (LLMs) using malicious or carefully crafted prompts have received increasing attention ~\cite{perez2022}. These techniques often referred to as prompt injection ~\cite{liu2024promptinjection,zhang2024promptinjection} or jailbreaking~\cite{chao2024jailbreaking,zhou2025tempest}, which are used to bypass safety guardrails or intentionally induce specific errors. These studies highlight inherent vulnerabilities in LLMs. Retrieval‑augmented generation (RAG) addresses part of this risk by querying external corpora and conditioning on retrieved evidence, thereby improving contextual relevance beyond what is stored in the model’s parameters and avoid hallucinations~\cite{li2024enhancingllmfactualaccuracy}. However, retrieval opens a complementary attack surface: adversaries can manipulate the ranking stage. We study this retrieval‑side vector via a black‑box, gradient‑free suffix that hijacks the evidence set, exposing the unsafe side of RAG.

\subsection{Evolutionary Optimization and Differential Evolution.}
Differential Evolution (DE)~\cite{DE1997} is a classic evolutionary algorithm for global optimization in continuous domains, noted for its simplicity and robustness. DE evolves a population of candidate solutions through repeated mutation (differential recombination of individuals) and crossover, selecting fitter candidates at each generation. It has been applied successfully in many black-box attack contexts~\cite{lin2020blackbox}. For example, Su \textit{et al.}~\cite{Su_2019} used DE to craft minimal adversarial perturbations on images (the "one-pixel attack"), demonstrating that DE can attack models where gradient methods falter or are not available. Our approach brings evolutionary search to the prompt optimization problem. We treat the selection of a sequence of discrete tokens as an optimization problem by defining a differentiable fitness function (based on retrieval similarity) and using DE to navigate the combinatorial space. We also draw inspiration from recent work on using evolutionary strategies for prompt generation, such as the method proposed by Liu et al.~\cite{guo2025evoprompt}. Although those focus on improving task performance rather than attacking, the concept could be mutual and alike. DE provides a strong foundation for black-box prompt attacks due to its ability to handle discrete, non-differentiable spaces and escape local optima that might trap gradient-based methods.

\subsection{Detection of Adversarial Prompts}
 Large Language Models (LLMs) become increasingly integrated into real-world applications recently, which ensures their robustness against adversarial prompts has become a critical research focus. Adversarial prompts refer to crafted inputs that exploit the system's vulnerabilities to elicit harmful, biased, or undesired outputs. To address this threat, various detection strategies have been proposed. Hu et al. (2023) ~\cite{hu2023token} introduced token-level adversarial prompt detection by analyzing the perplexity of each token in the input and visualizing suspicious regions through heatmaps, allowing granular inspection of adversarial content . Also, Kim et al.~\cite{kim2024aps} propose Adversarial Prompt Shield (APS), a lightweight safety classifier built on DistilBERT: the input prompt is tokenized and passed through the transformer, and the final-layer [CLS] embedding is fed into a binary head that labels each prompt for safe classification . Alon and Kamfonas~\cite{alon2023detecting} introduce a perplexity-based approach in which a reference language model computes token- or sequence-level perplexity; inputs whose perplexity exceeds a learned threshold—signaling the atypical linguistic patterns of adversarial suffixes—are flagged as attacks.

\subsection{Retrieval Mechanisms}
Sparse retrieval methods rely on exact or approximate term matching to efficiently filter large document collections. A classic example is BM25, introduced by Robertson and Zaragoza ~\cite{robertson2009probabilistic}, which computes query–document relevance scores using TF–IDF and document‐length normalization to achieve fast, keyword‐based retrieval. Dense retrieval approaches instead learn continuous vector representations for queries and passages. Karpukhin \emph{et al.} ~\cite{karpukhin2020dense} proposed Dense Passage Retrieval (DPR), a dual‐encoder framework that encodes queries and documents into a shared embedding space, significantly outperforming BM25 in top‐K retrieval accuracy on several open‐domain QA benchmarks.

\section{Prompt Adversarial Attacks}
We propose our prompt adversarial attack approach in this section. We frame the problem of finding a short adversarial suffix as a black-box optimization over discrete token sequences~\cite{diao2023blackbox}.  Rather than relying on gradients, we employ \emph{Differential Evolution} (DE) to evolve a population of candidate suffixes until one successfully steers the retriever to surface a chosen “target” document in its top-\(k\). First, we introduce our overall architecture, followed by an explanation of how Differential Evolution (DE) shifts the embedding vector within the representation space toward a desired target. The DE process is directly guided by the target corpus, which leads to strong performance.

\begin{figure}[!htbp]
  \centering
  \includegraphics[width=0.5\textwidth]{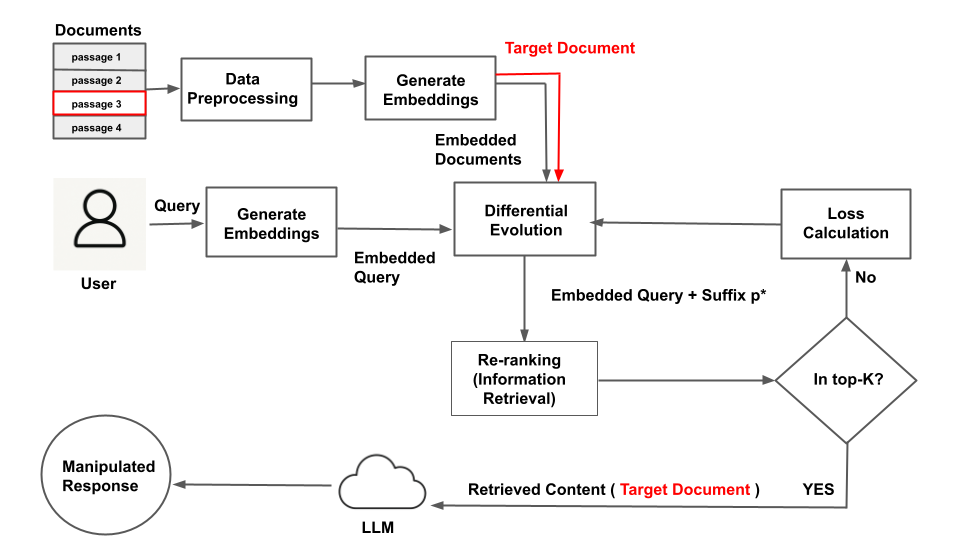}
  \caption{System architecture of the differential evolution (DE) adversarial attack in a RAG pipeline. The diagram shows how a population of candidate suffixes is initialized, evaluated ranking against the target document, and retrieved the target document resulted in manipulated response generated by Large Language Model.}
  \label{fig:de-architecture}
\end{figure}

Figure~\ref{fig:de-architecture} illustrates how the DE‐based workflow injects adversarial suffixes into the RAG pipeline, thereby paving the way for the attack methodology detailed in the following section.

\subsection{Theory Analysis}

\subsubsection{Local Sensitivity Surface and Attack Implications}
Smoothness in all directions does not rule out the existence of a single highly sensitive direction, to exploit the vulnerability of RAG based models,
we estimate the unit vector $\mathbf d_1$ that maximizes the directional derivative $\nabla_{\mathbf q}\cos(\mathbf q,\mathbf d)$
among random unit vectors $u$:
\begin{equation}
  \mathbf d_1
  = 
  \arg\max_{u:\,\lVert u\rVert=1}
  \frac{\cos(\mathbf q+\eta u,\mathbf d)-\cos(\mathbf q,\mathbf d)}{\eta},
  \quad\eta=10^{-3}
  \label{eq:d1}
\end{equation}
We then pick an orthogonal unit vector $\mathbf d_2$ and scan
$(\alpha,\beta)\!\in[-1,1]^2$:
\begin{equation}
  \mathbf q' = \mathbf q + \alpha\,\mathbf d_1 + \beta\,\mathbf d_2,
  \quad
  f(\alpha,\beta) = \cos(\mathbf q',\mathbf d).
  \label{eq:qprime}
\end{equation}
Figure~\ref{fig:local_surface} shows the 3-D surface result, where
$\alpha$ follows the \emph{steepest direction} and
$\beta$ lies in its orthogonal subspace.
The score increases almost monotonically along $\alpha$ while remaining flat
along $\beta$, clarifying the result that $\mathbf d_1$ dominates to this task.
Therefore, the next step is to find an algorithm that can highly adapt to the local direction. We then figure out Differential Evolution (DE), which is highly effective on this type of task. To prove the effectiveness of Differential Evolution, we designed an experiment, keeping the corpus size fixed at $|C|$ passages. For every query $\mathbf q$ we first rank $|C|$ passages with a dense retriever and
select the fixed rank negative passage $\mathbf t$ as the attack target. Using a fixed amount of tokens as suffix under DE to observe the effectiveness, we optimize the robust hinge loss
\begin{equation}
  L(\mathbf q) =
  \max_{1 \le i \le 12}
  \bigl[s_{(10)}^{(i)} - s_{\mathrm{tgt}}^{(i)}\bigr]_{+},
  \qquad
  \varepsilon_{\mathrm{noise}} = 0.2
  \label{eq:robust_hinge}
\end{equation}
where  Gaussian perturbations
$\delta_i\!\sim\!\mathcal N\!\bigl(0,\varepsilon^2\mathbf I\bigr)$ are
sampled per evaluation. The attack stops once the algorithm finds the answer at $L=0$.

For each query–target pair we further compute an isotropic local
slope at $\varepsilon=0.4$:
\begin{equation}
  \lambda(\mathbf q) =
  \frac{\bigl|\cos(\mathbf q+\delta,\mathbf t) - \cos(\mathbf q,\mathbf t)\bigr|}
       {\lVert\delta\rVert},
  \quad
  \delta \sim \mathcal{N}(0,\,0.4^2 \mathbf I)
  \label{eq:local_slope}
\end{equation}
to check whether it is still effective under noised situations. Our theoretical analysis result is given in Appendix~\ref{app:theory}.

\subsection{Problem Formulation}
We concentrate on the two predominant retrieval paradigms—sparse retrievers and dense retrievers—which together account for RAG systems. Since the procedure is analogous, we present the attack on dense retrievers and obtain the sparse variant by replacing the cosine similarity score with the BM25 score.



For dense retrieval, we encode queries and passages into a shared embedding space via a dual‐encoder and rank by cosine similarity.\\
Let a user query be \(q\) and a document corpus
\[
  X = \{X_1, \dots, X_N\}.
\]
A RAG retriever first embeds any text via
\[
  M:\text{text} \;\rightarrow\; \mathbb{R}^d,
\]
then ranks each \(X_i\) according to cosine similarity. Denote\vspace{-3pt}
\[
  e_q       = M(q), 
  \quad
  e_{q\Vert s} = M\bigl(q \Vert s\bigr), 
  \quad
  e_t       = M(X_t),
\]
where \(s\) is an adversarial \emph{suffix}, “\(\Vert\)” concatenates tokens, and \(X_t\) is the \emph{target} passage.

\paragraph{Ranking operator.}
Let $\smash{\tau_k(e)}$ be the $k$-th largest cosine similarity between a
query embedding $e$ and every document embedding
$\{M(X_i)\}_{i=1}^N$:\vspace{-4pt}
\begin{align}
  \tau_k(e) &=
  \text{$k$-th\ largest}\;
  \Bigl\{
    \mathrm{Sim}\bigl(e,M(X_i)\bigr)
  \Bigr\}_{i=1}^N,
  \label{eq:tau}
\end{align}
where $\mathrm{Sim}(u,v)=\tfrac{u\cdot v}{\lVert u\rVert\lVert v\rVert}$.
For succinctness we also write
$\operatorname{rank}(e,e_t)=
      \bigl|\{i:\mathrm{Sim}(e,M(X_i))>\mathrm{Sim}(e,e_t)\}\bigr|+1$.
\paragraph{Hinge loss.}
We measure the gap between the target and the current top-$k$ threshold through
\begin{equation}
  \label{eq:loss}
  L(s)
  =
  \max\bigl\{0,\;
    \tau_k\bigl(e_{q\Vert s}\bigr)
    -\mathrm{Sim}\bigl(e_{q\Vert s},e_t\bigr)
  \bigr\},
\end{equation}
so that
\begin{equation}
  \label{eq:rank-criterion}
  L(s) = 0
  \;\Longleftrightarrow\;
  \operatorname{rank}\bigl(e_{q\Vert s},e_t\bigr) \le k.
\end{equation}

Our attack therefore solves the discrete program
\begin{equation}
  \label{eq:opt}
  \min_{s\in\mathcal V^{\le n_{\max}}} L(s),
\end{equation}
where $\mathcal V$ is the vocabulary and $n_{\max}$ the suffix-length budget.
We then adapt the method to sparse retrievers by replacing cosine similarity with the BM25 score.

\subsection{Differential Evolution for Suffix Optimization}
\label{sec:DE}
Equation~\eqref{eq:opt} involves a black-box, non-differentiable objective function $L(s)$ (the black-box loss function that takes a full token sequence $s$ and returns a scalar fitness value). Therefore, we adopt the Differential Evolution (DE) algorithm~\cite{DE1997} to search for an optimal solution. Below, we describe the three main phases of DE—\emph{mutation}, \emph{crossover}, and \emph{selection}—and clarify every symbol and parameter used. 
Below we detail the workflow for generating a candidate suffix to a final adversarial suffix eventually.


\subsubsection{Encoding and Initialization}
Each candidate suffix (individual) is encoded as a fixed-length sequence of tokens of length $n_{\max}$ (fixed length of each individual’s token sequence, i.e., 1 to 10 in our case study). Denote the population size by $N$ (number of individuals per generation), and label the population as
\[
\mathcal{S} = \{\,s_1,\,s_2,\,\dots,\,s_N\}, 
\quad
s_i = \bigl[s_i^{(1)},\,s_i^{(2)},\,\dots,\,s_i^{(n_{\max})}\bigr],
\]
where $s_i^{(d)}$ is the token at position $d$ in individual $i$. The initial population $\mathcal{S}$ is generated randomly, subject to the constraint that each position $s_i^{(d)}$ must be a valid token from the allowed vocabulary.

\subsubsection{Mutation}
For each target individual $s_i$ in the current generation, randomly select three distinct “parent” individuals $s_a$, $s_b$, and $s_c$ from the set $\{s_1,\dots,s_N\}\setminus\{s_i\}$ (three distinct parent indices). Their indices $a,b,c$ are drawn uniformly at random from $\{1,\dots,N\}\setminus\{i\}$, with $a\neq b\neq c$. We then compute a \emph{donor vector} $m$ as
\[
m \;=\; s_a \;+\; F\,\bigl(s_b - s_c\bigr)\,,
\]
where $F > 0$ is the \emph{scale factor} (controls how strongly the difference $(s_b - s_c)$ influences $m$; typically chosen in $[0.5,\,1.0]$). Because our individuals are discrete token sequences, the implementation usually proceeds by first mapping each token to a continuous representation (e.g., an embedding or an integer index), performing the arithmetic operation to form $m = [\,m^{(1)},\dots,m^{(n_{\max})}\,]$, and then projecting or discretizing each $m^{(d)}$ back into the nearest valid token. In any case, after discretization, $m$ is a length-$n_{\max}$ token sequence lying within the allowed vocabulary.

\subsubsection{Crossover and Selection}
Next, we combine the donor vector $m$ with the original target vector $s_i$ to produce a trial vector $\tilde s_i$, and then select the better of $s_i$ and $\tilde s_i$ under the black-box loss $L(\cdot)$.

Let $\mathrm{CR}$ denote the crossover rate, i.e.\ the probability of inheriting a component from $m$ instead of $s_i$ (typical values lie in $[0.1,\,0.9]$). We perform a binomial crossover as follows:

For each position $d = 1,\dots,n_{\max}$, draw 
\[
r^{(d)} \sim \mathcal{U}(0,1),
\]
an independent uniform random variable that decides whether position $d$ in $\tilde s_i$ comes from $m$ or from $s_i$. To guarantee that $\tilde s_i$ differs from $s_i$ in at least one position, we also randomly choose a single index 
\[
d^\star \in \{1,2,\dots,n_{\max}\}.
\]
Then set
\[
\tilde s_i^{(d)} =
\begin{cases}
m^{(d)}, & \text{if } r^{(d)} < \mathrm{CR} \text{ or } d = d^\star,\\[6pt]
s_i^{(d)}, & \text{otherwise.}
\end{cases}
\]
In other words, for each $d$, with probability $\mathrm{CR}$ we inherit $m^{(d)}$; otherwise, we keep $s_i^{(d)}$. The special index $d = d^\star$ forces at least one token to come from $m$, ensuring $\tilde s_i \neq s_i$ in every generation. The resulting $\tilde s_i = [\,\tilde s_i^{(1)}, \dots, \tilde s_i^{(n_{\max})}\,]$ remains a valid length-$n_{\max}$ token sequence.

Finally, we compare the objective values of the original target $s_i$ and the trial vector $\tilde s_i$ under the black-box loss $L(\cdot)$ (the black-box loss function that takes a full token sequence $s$ and returns a scalar fitness value). The better (lower) loss is retained in the next generation. Formally:
\[
s_i^{\text{new}} \;=\; 
\begin{cases}
\tilde s_i, 
& \text{if } L\bigl(\tilde s_i\bigr) \,\le\, L\bigl(s_i\bigr), 
\\[6pt]
s_i, 
& \text{otherwise.}
\end{cases}
\]
That is, if $\tilde s_i$ yields a loss no worse than $s_i$, we replace $s_i$ with $\tilde s_i$; otherwise, we keep $s_i$ unchanged. Repeating this process for all $i = 1,\dots,N$ completes one generation of DE. We then iterate mutation, crossover, and selection until a stopping criterion is met (e.g., reaching a maximum number of generations).

Reiterating the process above, each generation of DE alternates between generating trial vectors via mutation and crossover, and then selecting the best candidates according to the loss function.In our implementation for adversarial suffix optimization, we additionally incorporate an early‐stop criterion based on a patience counter to halt each suffix stage once no improvement is observed for a predefined number of generations. Here, we use a suffix token sequence as an illustrative example to demonstrate how DE operates on discrete token sequences. To consolidate the description above, Algorithm~\ref{alg:de_with_stop} presents the complete pseudo code for the DE procedure with early stopping when optimizing a suffix token sequence.
\renewcommand{\KwRet}[1]{\textbf{return } #1\;}
\newcommand{\textproc}[1]{\textsc{#1}}

\begin{algorithm}[htbp]
\SetAlgoNlRelativeSize{-1}
\caption{DE with Early Stop for Optimizing a Suffix Token Sequence}
\label{alg:de_with_stop}
\DontPrintSemicolon
\KwIn{Population size $N$ (number of candidate suffixes); generations per stage $G$ (max iterations per suffix length);\\
      crossover rate $\mathrm{CR}$ (probability of crossover per token); mutation factor $\mathrm{F}$ (scaling for mutant vector);\\
      max suffix length $n_{\max}$; plateau patience $T$ (number of generations without improvement before early stop)}
\KwOut{Optimized adversarial suffix $s^\ast$ (token sequence)}
$s^\ast \gets []$ \tcp*{start with empty suffix}
\For{$L \gets 1$ \KwTo $n_{\max}$}{
  initialize population $\{s_i\}_{i=1}^{N}$ by right-padding $s^\ast$ to length $L$\;
  $\mathit{best}\gets\infty,\;\mathit{pat}\gets0$\;
  \For{$g \gets 1$ \KwTo $G$}{
    \For{$i \gets 1$ \KwTo $N$}{
      pick distinct indices $a,b,c\neq i$\;
      $m\gets\textproc{MutXover}(s_a,s_b,s_c,\mathrm{CR},F)$\;
      \If{$L(m)<L(s_i)$}{$s_i\gets m$}
    }
    $L_{\min}\gets \min_i L(s_i)$\;
    \eIf{$L_{\min}<\mathit{best}$}{
      $\mathit{best}\gets L_{\min},\;\mathit{pat}\gets0$\;
    }{
      $\mathit{pat}\gets \mathit{pat}+1$\;
    }
    \If{$\mathit{best}=0$ \textbf{or} $\mathit{pat}\ge T$}{\textbf{break}}
  }
  $s^\ast\gets \arg\min_i L(s_i)$\;
  \If{$L(s^\ast)=0$}{\textbf{break}}
}
\KwRet $s^\ast$
\end{algorithm}
Now, to illustrate suffix‐level DE in action, we give a brief example.
\noindent We initialize the example by taking the query \emph{“What is the capital of France?”} and documents X$_1$ (“Paris is the capital and most populous city of France.”), X$_2$ (“Berlin is the capital of Germany.”), and X$_3$ (“Madrid is the capital of Spain.”). Our goal is to promote X$_3$ by appending one token from 
\[
V = \{[\mathtt{unused186}],\,\mathtt{wash},\,\mathtt{candidate}\}.
\]
We run DE with \(n_{\max}=1\), \(N=3\), \(F=0.5\), \(\mathrm{CR}=0.5\), initializing suffixes \((s_1,s_2,s_3)=([\mathtt{unused186}],\;\mathtt{wash},\;\mathtt{candidate})\) and extracting their 768-dimensional embeddings using the BERT encoder. Next, we perform mutation by randomly selecting three parent tokens—\texttt{phantom}, \(\heartsuit\), and \texttt{token 1634}—from the full tokenizer list. 
Denoting their embeddings by \(e_{p_a},e_{p_b},e_{p_c}\), the first three dimensions are:
\[
\begin{aligned}
  e_{p_a}^{1\text{–}3} &= [-0.045,\,-0.080,\,-0.005],\\
  e_{p_b}^{1\text{–}3} &= [-0.011,\,-0.044,\,0.013],\\
  e_{p_c}^{1\text{–}3} &= [-0.022,\,-0.082,\,-0.010].
\end{aligned}
\]
We then compute the donor vector
\[
m = e_{p_a} + 0.5\,(e_{p_b}-e_{p_c}),
\]
yielding \(m^{1\text{–}3}=[-0.0398,\,-0.0619,\,0.0062]\). An \(L_2\) nearest-neighbor search over \(V\) selects \texttt{phantom}. Lastly, for the individual originally “wash,” a draw \(r=0.37<0.5\) yields trial \(\tilde s=\mathtt{phantom}\); computing hinge loss
\[
L(s)=\max\{0,\;\max_{j\neq3}\mathrm{sim}(q\!\parallel\!s,X_j)-\mathrm{sim}(q\!\parallel\!s,X_3)\},
\]
we obtain \(L(\mathtt{wash})\approx0.4\) versus \(L(\mathtt{phantom})\approx0.2\). Since the trial loss is lower, we update \(s_2\leftarrow\mathtt{phantom}\). Repeating until \(L(s)=0\) produces a suffix that guarantees X$_3$ becomes the top-1 result.



\textbf{Stopping criteria.} We combine two complementary rules:
\begin{itemize}[leftmargin=1.5em]
  \item \textbf{Success rule:}
        \(
          \mathbb{I}_{\text{succ}}(s)=
          \bigl[
            \operatorname{rank}(e_{q\Vert s},e_t)\!\le\!k
            \wedge
            \operatorname{rank}(e_{q\Vert s},e_q)\!>\!k
          \bigr].
        \)
        If \(\mathbb{I}_{\text{succ}}(s)=1\) for any population member,
        evolution halts immediately.
  \item \textbf{Plateau rule:}
        with \(B_g=\min_i L^{(g)}(s_i)\), stop when
        \(B_g-B_{g-T}=0\), i.e.\ no improvement for \(T\) consecutive
        generations.
\end{itemize}

Our DE framework includes three complementary variants that both rely on the same success and plateau rules but optimize for different objectives. The \textsc{DE\_fixed\_stop} variant applies these rules to a single, predetermined suffix length and is tuned for speed—it terminates as soon as the success rule is satisfied or the search stagnates, yielding a solution in the fewest iterations. In contrast, \textsc{DE\_seq\_stop} and \textsc{DE\_seq}  incrementally increases the suffix length in stages, invoking the stopping criteria at each stage in plateau rule and success rule individually; this approach may require more iterations but returns the shortest possible adversarial suffix, minimizing the number of tokens added. Section~\ref{sec:exp} shows that our hybrid early-stopping strategy cuts average query cost by $\approx40\%$ while matching the
attack success of vanilla DE.

\section{Evaluation}\label{sec:exp}
\subsection{Dataset and Setting}
To theoretically validate our attack effectiveness, we use the MS MARCO passage retrieval dataset~\cite{bajaj2018msmarcohumangenerated}. Then, we evaluate our attacks on  more BEIR benchmarks—\textbf{FiQA-2018} (financial QA)~\cite{Maia2018Fiqa}, \textbf{FEVER} (fact extraction and verification)~\cite{thorne2018fever}, and \textbf{SciFact} (scientific fact verification)~\cite{SciFact2020} to better compare to other existing methods. The BEIR framework~\cite{Thakur2021Beir} is used to standardize evaluation across these datasets. For each dataset, we randomly sample a subset of 1,000 documents and 100 queries from the official corpus and queries or claims splits.  Each query has one or more annotated relevant passages \(X_u\).  For the adversarial attack we additionally choose, for each query, a single “target” wrong passage \(X_t\) (either drawn uniformly at random from the non-relevant documents, or selected as a topically confusable distractor). Data and experiment results are available at \url{https://github.com/pen9rum/Rag_attack_DeRag}~~\cite{pen9rum2025RagAttackDeRag}.

For the dense retriever, all documents and queries are encoded with a BERT-base-uncased encoder ~\cite{devlin2018bert} (the same model used for adversarial patch generation) into 768-dimensional CLS embeddings, comprising over 110 million parameters. Retrieval is performed by computing cosine similarity over these embeddings and returning the top-\(k\) passages (with \(k\)=1, 10, 20). For sparse retrievers, we index the same corpus with the BM25 algorithm, which computes relevance scores using term frequency–inverse document frequency (TF–IDF) and document-length normalization, and likewise retrieve the top-\(k\) passages based on these BM25 scores.
\paragraph{Evaluation metrics.}
RAG pipelines can be subverted by adversarial prompts that promote an incorrect target passage \(X_t\) into early ranks. To evaluate both the efficacy and cost of such attacks over 100 queries, we employ six key measures.  We assess each method over 100 queries using a suite of retrieval‐oriented and optimization‐oriented measures.  First, \emph{Success@K} captures the proportion of queries for which the true target document \(X_t\) appears within the top-\(K\) retrieved results directly. \emph{Average tokens} records the mean length of the adversarial suffix, reflecting the minimal perturbation size required. \emph{Average iterations} counts the mean number of optimization steps , noticing that it reflects the actual number of iterations performed by the DE algorithm, rather than the hyperparameter‐defined iteration limit to capture computational effort. \(\Delta\mathrm{MRR}_t\) measures the change in reciprocal rank \(1/r\) of \(X_t\) before and after the attack, isolating the shift in ranking. \(\Delta\mathrm{nDCG@20}_t\) computes the difference in normalized discounted cumulative gain at cutoff 20 when treating \(X_t\) as the sole relevant item, assessing retrieval quality degradation.  \(\Delta\cos = \cos(e_{q\parallel s},e_t) - \cos(e_s,e_t)\) captures the shift in semantic similarity between the adversarially augmented query and the target; a smaller \(\lvert\Delta\cos\rvert\) indicates a subtler semantic perturbation, which is likely more difficult for detectors to identify. 


\subsection{Prompt perturbation results on Sparse Retriever Attack}
To clarify, our method is effective across multiple retrieval approaches, so we also conducted experiments on sparse retrieval.
In Table~\ref{tab:sparse-attack-result}, we compare our DE-based attacks against PRADA—a state-of-the-art sparse-retrieval black-box method ~\cite{wu2022prada}. Across four sparse‐retrieval benchmarks, our sequential variant \textsc{DE\_seq\_stop} achieves the highest Top-10 and Top-20 success rates while perturbing only an average of 2–3 tokens per query. By employing early stopping, the suffixes remain extremely compact without compromising their effectiveness. When minimizing query length is crucial, our fixed-length variant \textsc{DE\_fixed\_stop} matches or outperforms PRADA at all thresholds and converges in fewer iterations. These results show that our DE framework provides superior control over both suffix length and query cost under sparse retrieval.

\begin{table}[!htb]
  \centering
  \caption{Tail-Patch performance on four datasets
           (1\,000-document subset, 100 queries) on sparse retrievers.}
  \label{tab:sparse-attack-result}
  \small
  \setlength{\tabcolsep}{4pt}
  \resizebox{0.5\textwidth}{!}{%
  \begin{tabular}{%
      l  
      r  
      l  
      S[table-format=1.3]  
      S[table-format=1.3]  
      S[table-format=1.3]  
      S[table-format=1.3]  
      S[table-format=4.0]  
      S[table-format=1.3]  
      S[table-format=1.3]  
    }
    \toprule
    Dataset & {N}
    & Method
    & {Succ@1} & {Succ@10} & {Succ@20}
    & {Avg  Tok} & {Avg  Iter}
    & {$\Delta$MRR} & {$\Delta$nDCG} \\
    \midrule
    \multirow{3}{*}{SciFact}
      & \multirow{3}{*}{1000}
      & DE\_seq\_stop   & 0.250 & 0.890 & 0.970 & 3.133 & 3378 & 0.262 & 0.413 \\
      &                     & DE\_fixed\_stop & 0.190 & 0.850 & 0.930 & 5.000 & 1329 & 0.265 & 0.412 \\
      &                     & PRADA           & 0.010 & 0.980 & 1.000 & 2.253 & 1802 & 0.221 & 0.398 \\
    \midrule
    \multirow{3}{*}{FiQA}
      & \multirow{3}{*}{1000}
      & DE\_seq\_stop   & 0.270 & 0.810 & 0.890 & 3.387 & 3776 & 0.279 & 0.407 \\
      &                     & DE\_fixed\_stop & 0.190 & 0.710 & 0.850 & 5.000 & 1497 & 0.240 & 0.371 \\
      &                     & PRADA           & 0.030 & 0.990 & 1.000 & 2.163 & 1369 & 0.285 & 0.448 \\
    \midrule
    \multirow{3}{*}{FEVER}
      & \multirow{3}{*}{1000}
      & DE\_seq\_stop   & 0.620 & 0.960 & 0.970 & 2.697 & 2182 & 0.431 & 0.553 \\
      &                     & DE\_fixed\_stop & 0.550 & 0.930 & 0.960 & 5.000 &  975 & 0.422 & 0.546 \\
      &                     & PRADA           & 0.070 & 0.960 & 0.910 & 2.137 &  925 & 0.275 & 0.429 \\
    \midrule
    \multirow{3}{*}{MS MARCO}
      & \multirow{3}{*}{1000}
      & DE\_seq\_stop   & 0.710 & 0.990 & 0.990 & 2.323 & 1818 & 0.469 & 0.588 \\
      &                     & DE\_fixed\_stop & 0.600 & 0.980 & 0.980 & 5.000 &  829 & 0.438 & 0.564 \\
      &                     & PRADA           & 0.020 & 0.990 & 0.970 & 2.070 &  579 & 0.228 & 0.401 \\
    \bottomrule
  \end{tabular}%
  }
\end{table}

\subsection{Prompt-perturbation results on Dense Retriever Attack}
Dense retrievers are more robust than sparse retrievers because they encode both queries and documents into a continuous semantic space, rather than relying on exact term matches and term‐frequency signals. Therefore, we mainly focus on attack comparison in dense retrievers.   We compare our method against the gradient-guided white-box baseline \textsc{GGPP}~\cite{Hu2024GGPP} to demonstrate that effective dense‐retrieval attacks need not rely on white-box access.  
Table \ref{tab:beir-full} compares two strong differential-evolution variants—\textsc{DE\_seq\_stop} and \textsc{DE\_fixed\_stop}—against the
gradient-guided baseline \textsc{ggpp} and a random suffix on four BEIR
subsets.
For each method we report top-$K$ success rates
(\textit{Succ@$K$}) and five cost/quality metrics introduced in
Section 4.  In real Retrieval-Augmented Generation (RAG) pipelines
$k$ is typically between 5 and 20; Lewis et al.\,~\cite{lewis2021nlp}
evaluate with $k\!\in\!\{5,10\}$.  To cover edge cases we also include
$k\!=\!1$. Appendix C illustrates representative adversarial suffixes produced by our attacks.

\begin{table}[!htbp]
  \centering
  \caption{Tail-Patch performance on four datasets
           (1\,000-document subset, 100 queries) on dense retrievers.}
  \label{tab:beir-full}
  \small
  \setlength{\tabcolsep}{3.8pt}
  \resizebox{0.45\textwidth}{!}{%
  \begin{tabular}{%
      l  
      r  
      l  
      l  
      S[table-format=1.3]  
      S[table-format=1.3]  
      S[table-format=1.3]  
      S[table-format=1.2]  
      S[table-format=4.0]  
    }
    \toprule
    Dataset & {N} & Example & Method
            & {Succ@1} & {Succ@10} & {Succ@20}
            & {Avg  Tok} & {Avg  Iter} \\
    \midrule
    \multirow{4}{*}{SciFact}
      & \multirow{4}{*}{1000}
      & \multirow{4}{*}{C.1}
      & DE\_seq\_stop   & 0.198 & 0.573 & 0.739 & 2.34 & 2933 \\
      &                     &                     & DE\_fixed\_stop & 0.104 & 0.469 & 0.641 & 5.00 & 1256 \\
      &                     &                     & ggpp            & 0.146 & 0.458 & 0.565 & 5.00 &    \multicolumn{1}{c}{\textemdash} \\
      &                     &                     & Random          & 0.000 & 0.115 & 0.239 & 5.00 &   32 \\
    \midrule
    \multirow{4}{*}{FiQA}
      & \multirow{4}{*}{1000}
      & \multirow{4}{*}{C.2}
      & DE\_seq\_stop   & 0.126 & 0.520 & 0.546 & 2.63 & 3403 \\
      &                     &                     & DE\_fixed\_stop & 0.063 & 0.429 & 0.515 & 5.00 & 1107 \\
      &                     &                     & ggpp            & 0.105 & 0.480 & 0.546 & 5.00 &   \multicolumn{1}{c}{\textemdash} \\
      &                     &                     & Random          & 0.011 & 0.122 & 0.152 & 5.00 &   43 \\
    \midrule
    \multirow{4}{*}{FEVER}
      & \multirow{4}{*}{1000}
      & \multirow{4}{*}{C.3}
      & DE\_seq\_stop   & 0.185 & 0.515 & 0.643 & 2.76 & 3336 \\
      &                     &                     & DE\_fixed\_stop & 0.122 & 0.515 & 0.541 & 5.00 &  993 \\
      &                     &                     & ggpp            & 0.122 & 0.545 & 0.408 & 5.00 &   \multicolumn{1}{c}{\textemdash} \\
      &                     &                     & Random          & 0.010 & 0.152 & 0.235 & 5.00 &   44 \\
    \midrule
    \multirow{4}{*}{MS MARCO}
      & \multirow{4}{*}{1000}
      & \multirow{4}{*}{C.4}
      & DE\_seq\_stop   & 0.570 & 1.000 & 1.000 & 1.32 &  496 \\
      &                     &                     & DE\_fixed\_stop & 0.660 & 0.970 & 1.000 & 5.00 &  361 \\
      &                     &                     & ggpp            & 0.830 & 1.000 & 1.000 & 5.00 &   \multicolumn{1}{c}{\textemdash} \\
      &                     &                     & Random          & 0.340 & 0.760 & 0.710 & 5.00 &   20 \\
    \bottomrule
  \end{tabular}}
\end{table}

Table~\ref{tab:beir-full} reports the Tail-Patch attack success rates, token budgets, iteration counts, and quality‐shift metrics on four datasets. Table~\ref{tab:de-variants-merged} (in Appendix) summarizes the average $\Delta$MRR, $\Delta$nDCG and $\Delta\cos$ at cutoffs $K\in\{1,10,20\}$ in different contrast.

    Generally speaking, \textsc{DE\_seq\_stop} achieves the highest success rate at all datasets, especially reaching up to 0.739 on SciFact dataset while modifying only 2–3 tokens. Also, we notice that under MS MARCO dataset, the success rate is relatively higher than other datasets. The possible reason is that the dataset's development corpus is relatively small and highly redundant, so even very compact perturbations,can successfully attacked and perturbed the ranking result, resulting in strong performance for all methods with far fewer iterations and tokens.  Across all benchmarks, the DE‐based methods produce smaller average cosine shifts than \textsc{ggpp}, indicating reduced semantic drift in the adversarial suffix and having a relatively better results across four benchmarks.  

\subsubsection{Iterations–Tokens Trade‐off}
As shown in Table~\ref{tab:beir-full}, there is a clear trade‐off between the length of the adversarial suffix and the number of optimization iterations required by each method. The thousands of iterations in the table use a detailed metric—counting by treating every DE mechanism step as one iteration.  If we instead count a whole DE generation cycle as one step, each attack needs only dozens to hundreds of iterations.Thus, the effective computational cost is far smaller than the raw iteration counts might suggest. The sequential DE variant (\textsc{DE\_seq\_stop}) consistently finds very compact suffixes—on average only about 2–3 tokens long—but does so at the expense of a large number of iterations (for example, 2.34 tokens and 2 933 iterations on SciFact, and 1.32 tokens and 496 iterations on MS MARCO). By contrast, the fixed‐length DE variant (\textsc{DE\_fixed\_stop}), which always uses the full five‐token budget, roughly halves the iteration count (e.g., 5 tokens and 1 256 iterations on SciFact, and 5 tokens and 361 iterations on MS MARCO) while still achieving comparable success rates. Finally, the gradient‐guided baseline (\textsc{ggpp}) converges in only fewer gradient steps but likewise consumes the maximum five tokens. Consequently, if minimizing the visible perturbation (suffix length) is the primary goal, \textsc{DE\_seq\_stop} is the preferable choice; if reducing the number of model queries (iterations) is more important, \textsc{DE\_fixed\_stop} offers a better balance; and when gradient access is available, \textsc{ggpp} delivers the fastest convergence at the cost of the largest token footprint.Further details and the full data underlying the quality shift analysis are provided in Appendix~\ref{app:quality-shift}.
\subsubsection{Advantage of Early Stopping}
\begin{figure}[H]
  \centering
  \begin{subfigure}[t]{0.48\linewidth}
    \includegraphics[width=\linewidth]{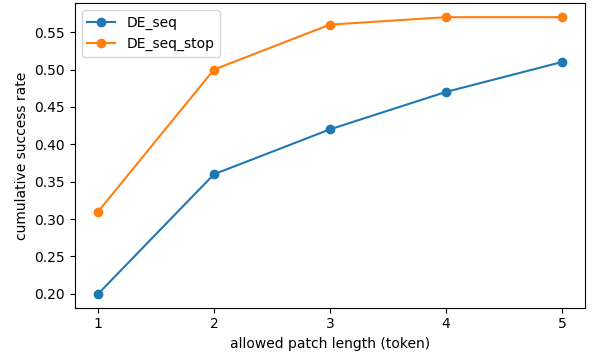}
    \caption{Cumulative success (Top-1).}
    \label{fig:cum-succ-1}
  \end{subfigure}\hfill
  \begin{subfigure}[t]{0.48\linewidth}
    \includegraphics[width=\linewidth]{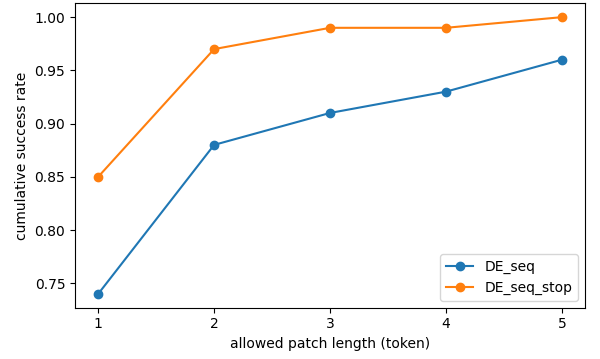}
    \caption{Cumulative success (Top-10).}
    \label{fig:cum-succ-10}
  \end{subfigure}
  \caption{Cumulative success rates for \texttt{DE\_seq} (blue) vs.\ \texttt{DE\_seq\_stop} (orange) as a function of allowed suffix length on MS MARCO dataset.}
  \label{fig:cum-success}
\end{figure}

Figure~\ref{fig:cum-success} clearly shows that the early‐stopping variant (\textsc{DE\_seq\_stop}) outperforms the vanilla progressive DE (\textsc{DE\_seq}) at every suffix length.  At Top-1 (Fig.~\ref{fig:cum-succ-1}), \textsc{DE\_seq\_stop} reaches a 50\% success rate with only 2 tokens, whereas \textsc{DE\_seq} achieves just 36\% at that length and only approaches 51\% by 5 tokens.  At Top-10 (Fig.~\ref{fig:cum-succ-10}), the early-stopping method attains 97\% success by 2 tokens and saturates above 99\% by 3 tokens, while \textsc{DE\_seq} needs 3–4 tokens to exceed 90\%.  These results demonstrate that early stopping not only increases cumulative success for a fixed budget but also identifies the minimal suffix length required—thereby saving unnecessary queries and limiting token perturbations.  

\subsubsection{Prefix and Suffix mutual Attacks}
We additionally ran DE on both prefix and suffix from 1 to 5 tokens, attacking on four datasets to examine whether the two strategies are complementary or not. To consider the best potential, we set the task on targeting rank 1.
Table~\ref{tab:prefix_suffix} reports the number of queries out of $100$ that each strategy succeeds .

\begin{table}[!htbp]
  \centering
  \caption{Prefix–suffix complementarity under \textsc{success@1}.}
  \resizebox{\columnwidth}{!}{%
  \small
  \begin{tabular}{lcccccc}
    \toprule
    Dataset &
    Suffix &
    Prefix&
    Both &
    Both &
    Comp.\@ &
    Either \\
    &
    only &
    only &
    succeed &
    fail &
    rate &
    succeed \\
    \midrule
    FIQA     & 14 & 14 & 17 & 55 & 28 & 45 \\
    FEVER    &  5 &  7 & 24 & 64 & 12 & 36 \\
    SciFact  & 11 & 15 & 15 & 59 & 26 & 41 \\
    \bottomrule
  \end{tabular}%
  }
  \label{tab:prefix_suffix}
\end{table}
We further tested a monotonic suffix length schedule and a naive cosine‑similarity objective; neither improved Success@K nor ranking displacement (Appendix~\ref{app:suf_len} and ~\ref{app:tab:loss_aggregate_comparison}).
These results show that if we want to achieve more successes, it is more efficient to insert tokens at different positions. Positional diversity in token-level attacks is valuable and gives great performance boost on generating adversarial prompt.

\subsection{Adversarial Query Crafting: Detection Bypass and Readability}
\label{sec:anti-detect}
\subsubsection{Evasion Performance against a BERT-Based Detector}
The numerical detector scores, along with the corresponding distributions, are deferred to Appendix~\ref{appendix:distributions}. Specifically, Table~\ref{tab:de-det-stats} summarizes the perplexity and BERT-CLS scores, while Fig.~\ref{fig:ppl-dist} and Fig.~\ref{fig:cls-dist} visualize their distributions respectively, where PPL stands for perplexity. On average, the PPL scores are higher in two datasets and relatively lower in the other two. Notably, the MS MARCO dataset contains an extreme outlier query, which significantly elevates its overall PPL score. In contrast, the BERT-CLS–based attack probability remains consistently similar across datasets, suggesting that, although manual inspection may still reveal differences between original and adversarial queries, these perturbations can deceive certain detection models. Future research should develop targeted attacks against emerging defense mechanisms.

We further evaluate our adversarial examples against the RoBERTa detector~\cite{Solaiman2019} to measure the detection rate. In total, we collected 783 samples (483 adversarial and 300 original clean queries) and assessed the detector’s ability to distinguish between them. Table~\ref{tab:detector_performance} summarizes detection metrics under various FPR targets.

\begin{table}[!htbp]
  \centering
  \caption{Detection performance of RoBERTa detector under different threshold.}
  \resizebox{\columnwidth}{!}{%
  \small
  \begin{tabular}{lccccc}
    \toprule
    \textbf{Target FPR} & \textbf{Actual FPR} & \textbf{Threshold} & \textbf{Precision} & \textbf{Recall} & \textbf{F1} \\
    \midrule
    0.5\%  & 0.33\% & 0.952 & 0.00\%  & 0.00\% & 0.00\%  \\
    1.0\%  & 0.33\% & 0.952 & 0.00\%  & 0.00\% & 0.00\%  \\
    2.0\%  & 2.00\% & 0.874 & 33.33\% & 0.62\% & 1.22\% \\
    \bottomrule
  \end{tabular}%
  }
  \label{tab:detector_performance}
\end{table}

The overall detection performance is poor: at a standard target FPR of 0.5\%, the detector achieves an AUROC of 0.2023 and an AUPRC of 0.4665. This indicates that prompt‐injection examples generated by DE attacks are nearly indistinguishable from benign inputs since it only required few tokens, therefore demonstrates that our token‐level perturbations effectively evade detection even at stringent false‐positive rates, providing strong evidence that DE prompt injection attacks can deceive detectors and are difficult to distinguish.
\label{subsec:attack_quality}
\begin{table*}[!htbp]
\centering
\caption{Answer quality degradation under adversarial retrieval. “Top-10 only” means target chunk elevated into Top-10 but not rank 1. “Fail” = target chunk not inserted into Top-10. “Average” is the weighted per-query best attacked outcome (Top-1 $\rightarrow$ Top-10 only $\rightarrow$ Fail) using the empirical rates. Percent changes are relative to each dataset's baseline.}
\label{tab:adv_one_block}
\small
\setlength{\tabcolsep}{7pt}
\begin{tabular}{l l c r r r r}
\toprule
Dataset & Group & Rate & EM & F1 & ROUGE-L & BERTScore \\
\midrule
\multirow{5}{*}{SQuAD (500)} 
  & \textbf{Baseline} & --    & \textbf{0.322} & \textbf{0.388} & \textbf{0.396} & \textbf{0.900} \\
  & Top-1             & 11.4\% & 0.053 (↓83.5\%) & 0.056 (↓85.6\%) & 0.056 (↓85.9\%) & 0.816 (↓9.3\%) \\
  & Top-10 only       & 42.0\% & 0.162 (↓49.7\%) & 0.190 (↓51.0\%) & 0.191 (↓51.8\%) & 0.860 (↓4.4\%) \\
  & Fail              & 46.6\% & 0.348 (↑8.1\%)  & 0.415 (↑7.0\%)  & 0.430 (↑8.6\%)  & 0.897 (↓0.3\%) \\
  & \textbf{Average}  & 100\%  & 0.236 (↓26.7\%) & 0.279 (↓28.1\%) & 0.287 (↓27.5\%) & 0.872 (↓3.1\%) \\
\midrule
\multirow{5}{*}{NQ-Open (500)} 
  & \textbf{Baseline} & --    & \textbf{0.782}  & \textbf{0.8236} & \textbf{0.8616} & \textbf{0.9603} \\
  & Top-1             & 8.4\%  & 0.6190 (↓20.8\%) & 0.6565 (↓20.3\%) & 0.6520 (↓24.3\%) & 0.9218 (↓4.0\%) \\
  & Top-10 only       & 91.4\% & 0.6696 (↓14.4\%) & 0.7117 (↓13.6\%) & 0.7485 (↓13.1\%) & 0.9365 (↓2.5\%) \\
  & Fail              & 0.2\%  & 1.0000 (↑27.9\%) & 1.0000 (↑21.4\%) & 1.0000 (↑16.1\%) & 1.0000 (↑4.1\%) \\
  & \textbf{Average}  & 100\%  & 0.6660 (↓14.8\%) & 0.7076 (↓14.1\%) & 0.7409 (↓14.0\%) & 0.9354 (↓2.6\%) \\
\bottomrule
\end{tabular}
\end{table*}
\subsubsection{Human Reading Availability Strategy}

To enhance human readability without degrading attack success rates, we propose a dynamic candidate‐pool construction strategy. For each query, we first encode it with BERT on CPU to obtain its CLS vector and compute cosine similarities against a pre-encoded matrix of document CLS vectors; the top \(N\) most similar documents are used solely to form the contrastive pool in our differential-evolution attack. Independently, to construct the token candidate pool, we mask the last \(\mathrm{TAIL\_L}\) tokens of the query, pass the masked sequence through the MLM head, average the predicted softmax probabilities over those \(\mathrm{TAIL\_L}\) positions, and select the top \(k\) vocabulary tokens as mask-fill candidates. 
We perform an ablation over multiple pooling size shown below. Table~\ref{tab:rank_mrr} shows that retrieval effectiveness ($K=1$ and $\Delta$MRR) remains stable across settings, and Table~\ref{tab:time_metrics} (Appendix G) shows that pool construction, query optimization, and overall attack time remain low.

\begin{table}[ht]
\centering
\caption{Rank@1 success rate and MRR under various pool sizes.}
\begin{tabular}{c|cc|cc|cc}
\hline
\bfseries Pool size & \multicolumn{2}{c|}{\bfseries Fever} & \multicolumn{2}{c|}{\bfseries FiQA} & \multicolumn{2}{c}{\bfseries SciFact} \\
& $K=1$ & $\Delta$MRR & $K=1$ & $\Delta$MRR & $K=1$ & $\Delta$MRR \\
\hline
500     & 0.19 & 0.277 & 0.24 & 0.345 & 0.24 & 0.348 \\
1\,000  & 0.19 & 0.277 & 0.28 & 0.373 & 0.26 & 0.384 \\
2\,000  & 0.22 & 0.307 & 0.24 & 0.358 & 0.22 & 0.358 \\
5\,000  & 0.24 & 0.345 & 0.33 & 0.420 & 0.30 & 0.411 \\
10\,000 & 0.34 & 0.424 & 0.34 & 0.438 & 0.33 & 0.427 \\
20\,000 & 0.38 & 0.457 & 0.27 & 0.377 & 0.27 & 0.379 \\
30\,522 & 0.29 & 0.391 & 0.32 & 0.412 & 0.25 & 0.351 \\
\hline
\end{tabular}
\label{tab:rank_mrr}
\end{table}

To quantify readability gains, we report the average MLM negative log‑likelihood (NLL), a proxy for fluency, in Appendix~\ref{app:readability}.

\subsection{Downstream Answer Quality Degradation Evaluation}
We evaluate a two‑phase adversarial suffix attack on a dense‑retrieval RAG pipeline to measure degradation of downstream answer quality. The attack is staged: first we drive a chosen (irrelevant) target chunk into the Top‑10 so that it can influence the generator’s context window; second, conditional on that success, we further optimize to promote the target to rank 1. We select two QA benchmarks: SQuAD ~\cite{rajpurkar2016squad} and NQ‑Open (500 training queries). SQuAD supplies short span‑style factoid answers drawn from comparatively tight passages, and NQ‑Open contains broader and noisier web‑sourced evidence, probing robustness under a larger semantic space. For each query we first run a baseline generation using the original Top‑10 retrieved chunks. We then optimize a 5‑token adversarial suffix (Differential Evolution, up to 120 iterations) to insert an unrelated target chunk into the Top‑10 and, if successful, to push it to rank 1. Outcomes are stratified into Top‑1 success, Top‑10‑only success, and Fail , which stands for the target never reaches the Top‑10. We then compare answer quality across these outcome strata to quantify semantic degradation.

Table~\ref{tab:adv_one_block} shows persistent degradation once an adversarial chunk occupies retrieval slots. On SQuAD, a substantial portion of attempts did not elevate the target all the way to rank 1 because we allotted fewer optimization iterations, yet the successful insertions still drove large semantic quality losses. On Natural Questions we used more iterations, achieved higher insertion success, and again observed pronounced declines in answer fidelity. The pattern indicates that even realistic, heterogeneous questions are vulnerable and that reliance on protecting only the very top rank is inadequate; limiting adversarial occupancy at any position within the narrow retrieval window is essential to preserve downstream answer quality. 
\section{Conclusion}
We introduce DeRAG, a novel black-box adversarial attack framework targeting Retrieval-Augmented Generation (RAG) systems via prompt suffix perturbations. DeRAG formulates the attack as a discrete optimization problem and leverages Differential Evolution with custom stopping rules to efficiently discover suffixes that misdirect retrieval to incorrect documents. Experiments show that DeRAG matches or even outperforms gradient-based and other existing methods, while using fewer tokens and inducing smaller embedding shifts across both dense and sparse retrievers, all while maintaining a moderate level of anti-detection ability.
We highlight the importance of factors like population diversity and suffix length, and validate early-stopping strategies that cut query costs without reducing effectiveness. We further evaluate the downstream results in real-world scenarios. DeRAG not only reveals critical RAG vulnerabilities but also informs future defenses such as prompt precision, embedding regularization, and anomaly detection—ultimately advancing robust, trustworthy AI.

\clearpage
\let\balance\relax          
\onecolumn                  
\appendix
\section{Effectiveness on Differential Evolution attacks}
\label{app:theory}
\begin{figure*}[!htbp]
  \centering
  \begin{subfigure}[b]{0.24\linewidth}
    \centering
    \includegraphics[width=\linewidth]{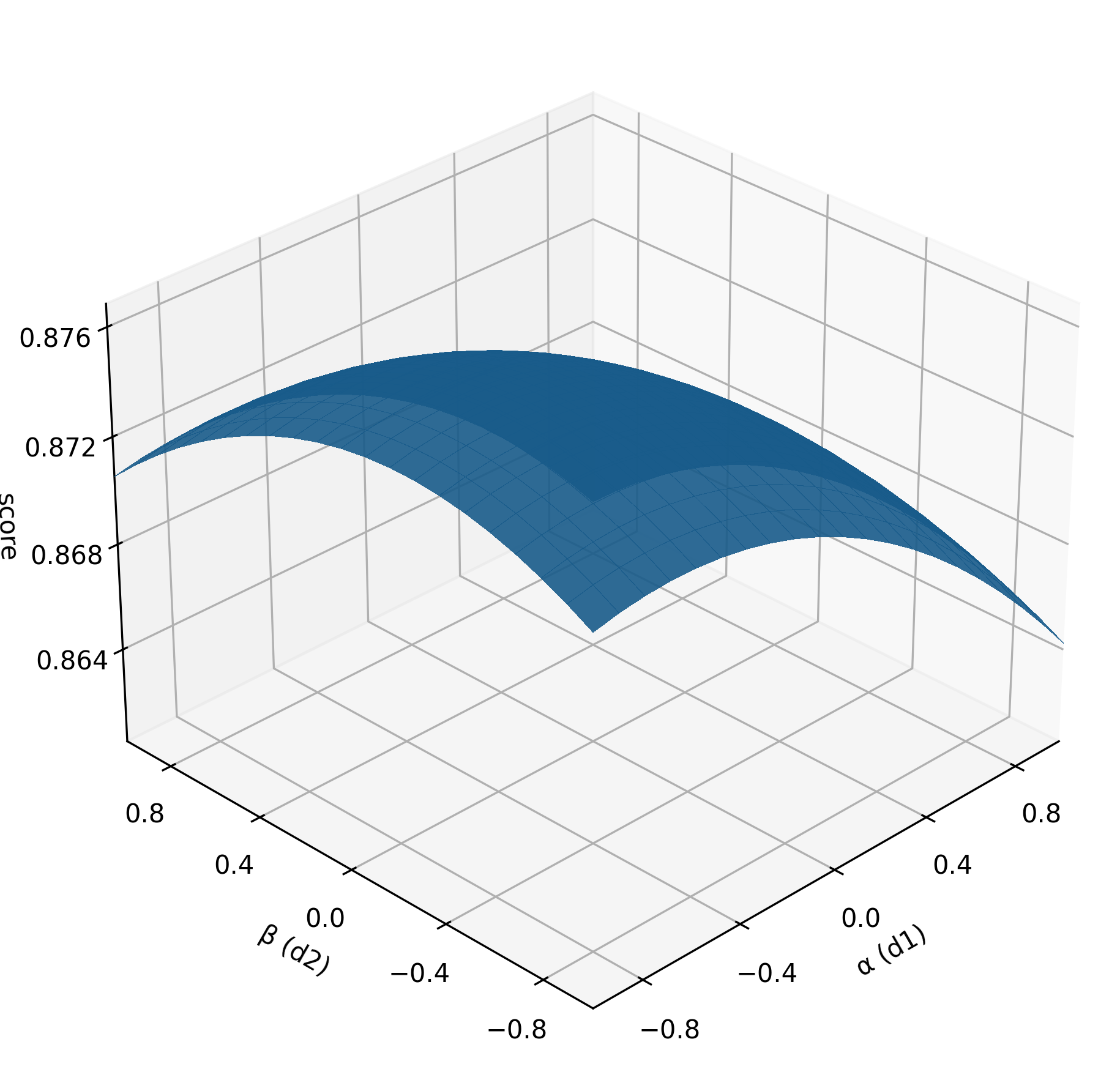}
    \caption{Local score surface around a query vector.}
    \label{fig:local_surface}
  \end{subfigure}\hfill
  \begin{subfigure}[b]{0.24\linewidth}
    \centering
    \includegraphics[width=\linewidth]{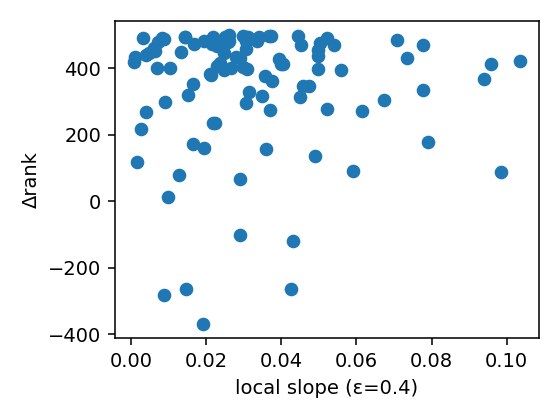}
    \caption{$\Delta$rank vs.\ local slope ($\varepsilon=0.4$)}
    \label{fig:slope_vs_rank}
  \end{subfigure}\hfill
  \begin{subfigure}[b]{0.24\linewidth}
    \centering
    \includegraphics[width=\linewidth]{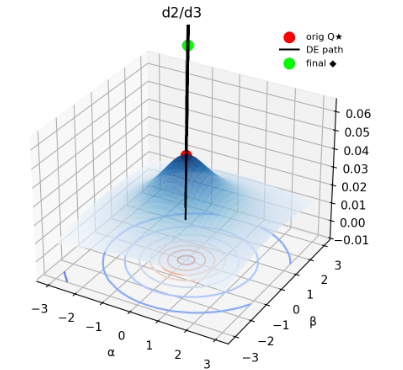}
    \caption{$(\mathbf d_2,\mathbf d_3)$ plane}
    \label{fig:surf_d2d3}
  \end{subfigure}\hfill
  \begin{subfigure}[b]{0.24\linewidth}
    \centering
    \includegraphics[width=\linewidth]{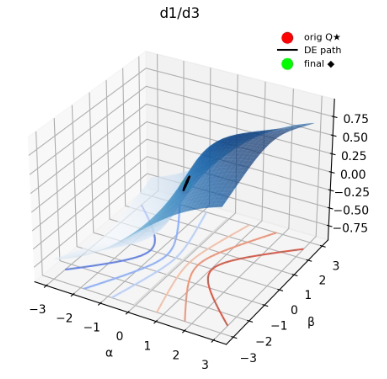}
    \caption{$(\mathbf d_1,\mathbf d_3)$ plane}
    \label{fig:surf_d1d3}
  \end{subfigure}

  \caption{%
    (a) Local score surface around a query vector.
    (b) Isotropic smoothness is nearly uncorrelated with attack efficacy.
    (c) Local score surface on the $(\mathbf d_2,\mathbf d_3)$ plane, emphasizing monotonicity along $\mathbf d_1$.
    (d) Local score surface on the $(\mathbf d_1,\mathbf d_3)$ plane.}
  \label{fig:combined_four}
\end{figure*}

Figure~\ref{fig:combined_four} shows the results. Over 100 queries under the condition of $|C|= 1000$ passages and twelve Gaussian perturbations, the mean slope is only 0.0328. Figure~\subref{fig:slope_vs_rank} plots $\Delta\mathrm{rank}$ against the local slope $\lambda(\mathbf q)$, yielding a Pearson correlation of $r=0.016$ (two‐tailed $p=0.871$),  proves that isotropic smoothness provides virtually no insight into how effectively an adversary can alter the document’s rank.
We then project the score of a representative query onto three orthogonal
planes using $\mathbf d_1$, $\mathbf d_2$, and $\mathbf d_3$.  
Figures ~\subref{fig:surf_d2d3} and \subref{fig:surf_d1d3} confirm that the score varies essentially only along $\mathbf d_1$, and the DE trajectory (black) climbs the ridge from the red star (original query) to the green diamond (final adversarial query), showing that DE exploits this structure with few tokens. Hence, retrievers remain vulnerable to concise suffix attacks.
\section{Quality Shift Result}\label{app:quality-shift}
Table~\ref{tab:de-variants-merged} compares three attacks (\textsc{DE\_seq\_stop}, \textsc{DE\_fixed\_stop}, \textsc{ggpp}) across
Top-$K$ settings ($K\in\{1,10,20\}$) using $\Delta$nDCG and $\Delta$MRR drops and cosine shift $\Delta\cos$ as our evaluation metrics.
At $K=1$, \textsc{ggpp} often attains the largest immediate ranking loss , but shows a larger (often positive) $\Delta\cos$, whereas \textsc{DE\_seq\_stop} attains comparable $\Delta$nDCG/$\Delta$MRR with a near‑zero shift, indicating a much lower semantic perturbation.
At $K=10$ and $20$, under more general (larger $K$) settings, \textsc{DE\_seq\_stop} surpasses the gradient‐guided \textsc{ggpp} (and the fixed variant) by matching or improving $\Delta$nDCG/$\Delta$MRR while keeping $\Delta\cos$ closest to zero.
Therefore, \textsc{DE\_seq\_stop} delivers the best balance—maintaining low semantic movement while producing equal or greater retrieval quality loss—making it the most effective and stealthy of the compared attacks.
\begin{table}[ht]
  \centering
  \caption{Quality shift $(\Delta)$ comparison of
           \textsc{DE\_seq\_stop}, \textsc{Fixed}, and \textsc{GGPP}
           at $K\!\in\!\{1,10,20\}$ on dense retriever.}
  \label{tab:de-variants-merged}
  \small
  \setlength{\tabcolsep}{4pt}
  \begin{tabular}{l l
                  S[table-format=1.3] S[table-format=1.3] S[table-format=1.3]
                  S[table-format=1.3] S[table-format=1.3] S[table-format=1.3]
                  S[table-format=1.3] S[table-format=1.3] S[table-format=1.3]}
    \toprule
    \multirow{2}{*}{Dataset} & \multirow{2}{*}{Metric}
      & \multicolumn{3}{c}{$K=1$}
      & \multicolumn{3}{c}{$K=10$}
      & \multicolumn{3}{c}{$K=20$} \\
   \cmidrule(lr){3-5}\cmidrule(lr){6-8}\cmidrule(l){9-11}
      &
      & \multicolumn{1}{c}{Seq}
      & \multicolumn{1}{c}{Fixed}
      & \multicolumn{1}{c}{GGPP}
      & \multicolumn{1}{c}{Seq}
      & \multicolumn{1}{c}{Fixed}
      & \multicolumn{1}{c}{GGPP}
      & \multicolumn{1}{c}{Seq}
      & \multicolumn{1}{c}{Fixed}
      & \multicolumn{1}{c}{GGPP} \\
    \midrule
    \multirow{3}{*}{SciFact}
      & $\Delta$nDCG & 0.188 & 0.267 & 0.226 & 0.196 & 0.212 & 0.203 & 0.198 & 0.167 & 0.153 \\
      & $\Delta$MRR  & 0.193 & 0.212 & 0.172 & 0.090 & 0.115 & 0.112 & 0.068 & 0.060 & 0.060 \\
      & $\Delta\cos$ & 0.002 & 0.038 & -0.004 & -0.008 & -0.006 & 0.031 & -0.006 & -0.039 & 0.011 \\[2pt]
    \multirow{3}{*}{FiQA}
      & $\Delta$nDCG & 0.113 & 0.201 & 0.147 & 0.186 & 0.176 & 0.201 & 0.141 & 0.145 & 0.158 \\
      & $\Delta$MRR  & 0.114 & 0.158 & 0.108 & 0.092 & 0.082 & 0.101 & 0.043 & 0.061 & 0.068 \\
      & $\Delta\cos$ & -0.003 & 0.008 & -0.019 & -0.010 & -0.016 & 0.024 & -0.015 & -0.012 & 0.020 \\[2pt]
    \multirow{3}{*}{FEVER}
      & $\Delta$nDCG & 0.181 & 0.203 & 0.237 & 0.110 & 0.110 & 0.140 & 0.181 & 0.176 & 0.140 \\
      & $\Delta$MRR  & 0.181 & 0.168 & 0.183 & 0.057 & 0.049 & 0.090 & 0.093 & 0.098 & 0.090 \\
      & $\Delta\cos$ & 0.021 & 0.111 & 0.090 & -0.005 & -0.019 & 0.084 & 0.040 & 0.060 & 0.084 \\[2pt]
    \multirow{3}{*}{MS MARCO}
      & $\Delta$nDCG & 0.722 & 0.646 & 0.585 & 0.597 & 0.646 & 0.607 & 0.474 & 0.555 & 0.607 \\
      & $\Delta$MRR  & 0.699 & 0.597 & 0.534 & 0.532 & 0.597 & 0.563 & 0.386 & 0.494 & 0.563 \\
      & $\Delta\cos$ & 0.015 & -0.027 & -0.021 & -0.052 & -0.027 & 0.016 & -0.048 & -0.016 & 0.016 \\
    \bottomrule
  \end{tabular}
\end{table}

\section{Adversarial Suffix Examples (Differential Evolution)}\label{app:adv-suffix}

\subsection{Examples on FiQA Dataset}\label{app:appendix:A1}
\begingroup
\setlength{\textfloatsep}{3pt}\setlength{\intextsep}{3pt}\setlength{\floatsep}{3pt}
\begin{table}[H]
  \centering
  \caption{Example DE prefix using user query and passages from the FiQA dataset, including suffix, original searched document, and retrieved (targeted) document.}
  \label{tab:FiQA-examples}
  \begin{tabular}{|c|p{3cm}|p{2.5cm}|p{5cm}|p{5cm}|}
    \hline
    \textbf{Ex \#} & \textbf{QUERY} & \textbf{Adversarial suffix} & \textbf{Original searched document} & \textbf{Retrieved (targeted) document} \\
    \hline
    ex1 & Historical stock prices: Where to find free / low cost data for offline analysis? 
        & \#\#systems williams truss ram exposing 
        & Go to \texttt{http://finance.google.com}, search for the stock you want. When you are seeing the stock information, in the top legend you can click “Download to CSV”. 
        & It’s a good point and I think you answered the question yourself: most IPOs don’t have that download option yet. There’s a real risk of stale data if you rely only on free sources. \\
    \hline
    ex2 & How can all these countries owe so much money? Why \& where did they borrow it from? 
        & \#\#‒ shannon wed 1995 liquid 
        & “They borrowed it from the people, and typically to finance wars and military spending. For example, Wikipedia suggests that many Eurozone governments issue bonds directly into primary markets.” 
        & “Never understood why any country would want to be beholden to complete strangers that you would never hire to begin with. Government debt usually goes to banks, but they resell it into the secondary market.” \\
    \hline
    ex3 & Why are auto leases stubbornly strict about visa status and how to work around that? 
        & \texttt{copyright advocacy affordable wu applied}
        & “Uh, you want to lease a car through a dealer? That is the worst possible way to obtain a car. Dealers love leases because it locks you into a contract. Try a local credit union—they often just need proof of residency, not citizenship.” 
        & “Also, the second amendment. Even if some kind of gun control measure were passed, sport shooting is never going to disappear—most people lease cars based on credit history, not visa status.” \\
    \hline
    ex4 & Should I re-allocate my portfolio now or let it balance out over time? 
        & solving \#\#™ merry \#\#nett livery 
        & “As you note, your question is inherently opinion-based. That said, if I were in your situation I would sell the stock that has a P/E above 30 and buy more of the index fund that tracks the S\&P 500.” 
        & “Text blast is the new marketing. I think any company should try it right away. It is cheap and effective. Texting services cost \$0.01 per SMS internationally but they boost open rates.” \\
    \hline
    ex5 & To pay off a student loan, should I save up a lump sum payoff payment or pay extra each month? 
       & \texttt{a 331 cameroon snorted rosenberg}
        & “As a new graduate, aside from the fact that you seem to have the extra \$193/mo to pay more towards your loan, we don’t know if you’ll keep that job. If your interest rate is 6.8\%, any extra payment cuts principal every month, so pay extra monthly if you can.” 
        & “It wasn’t a small hole in your suggestion. You were proposing a major law which already exists. The only anti-intellectual property you need is ‘First sale doctrine’. It already covers reselling.” \\
    \hline
  \end{tabular}
\end{table}
\endgroup

\subsection{Examples on SciFact Dataset}\label{app:appendix:A2}
\begingroup
\setlength{\textfloatsep}{3pt}\setlength{\intextsep}{3pt}\setlength{\floatsep}{3pt}
\begin{table}[H]
  \centering
  \caption{Example DE prefix using user query and passages from the SciFact dataset, including suffix, original searched document, and retrieved (targeted) document.}
  \label{tab:SCIFACT-examples}
  \begin{tabular}{|c|p{3cm}|p{2.5cm}|p{5cm}|p{5cm}|}
    \hline
    \textbf{Ex \#} & \textbf{QUERY} & \textbf{Adversarial suffix} & \textbf{Original searched document} & \textbf{Retrieved (targeted) document} \\
    \hline
    ex1 & The density of cytokine receptor bearing cells has no effect on the distance over which cytokines act. 
        & converse behaviour excitedly syllables homeland 
        & Immune cells communicate by exchanging cytokines to achieve a context‐appropriate response, but the distances over which those cytokines diffuse remain disputed. 
        & Neutrophil extracellular traps (NETs) are made of processed chromatin bound to granular and selected cytoplasmic protein, and these NETs can trap pathogens outside the cell. \\
    \hline
    ex2 & CCL19 is a ligand for CCR7. 
        & \texttt{baltic chang \#\#chus \#\#ead wnba}
        & Chemokines orchestrate immune cell trafficking by eliciting either directed or random migration and by activating integrins on leukocytes. 
        & TNF expression of macrophages is under stringent translational control that depends on the p38 MAPK/MK2 pathway and the eIF4E phosphorylation. \\
    \hline
    ex3 & A deficiency of vitamin B6 decreases blood levels of homocysteine. 
        & \texttt{\#\#z bal vocalist \#\#san eager}
        & BACKGROUND Homocysteine is a risk factor for coronary artery disease (CAD), although a causal relation remains to be proven. 
        & Introduction: Among the inflammatory mediators involved in the pathogenesis of obesity, the cell adhesion molecules P-selectin and ICAM-1 are up-regulated. \\
    \hline
    ex4 & Healthy volunteers exhibit rapid and transient increase of cellular ATP after being bolus-injected with fructose. 
        & evangelist tumbling rodeo assignment beard 
        & CONTEXT The mechanisms that drive progression from fatty liver to steatohepatitis and cirrhosis are unknown. In animal models, fructose injection can cause acute ATP depletion in hepatocytes. 
        & We investigated whether or not administered leptin influences anxiety-like behavior in ob/ob mice. Repeated intraperitoneal injection elevated hippocampal CREB phosphorylation. \\
    \hline
    ex5 & Gastric infection with \textit{Helicobacter pylori} decreases risk of gastric cancer in humans. 
        & reporter nicaragua cmll sgt \#\#idae 
        & OBJECTIVE To investigate the association between gastric cancer and prior infection with \textit{H. pylori}. DESIGN Case-control study in high-incidence region. 
        & Neutrophil extracellular traps (NETs) are made of processed chromatin bound to granular and selected cytoplasmic protein, which may contribute to tissue damage during sepsis. \\
    \hline
  \end{tabular}
\end{table}
\endgroup

\subsection{Examples on FEVER Dataset}\label{app:appendix:A3}
\begingroup
\setlength{\textfloatsep}{3pt}\setlength{\intextsep}{3pt}\setlength{\floatsep}{3pt}
\begin{table}[H]
  \centering
  \caption{Example DE prefix using user query and passages from the FEVER dataset, including suffix, original searched document, and retrieved (targeted) document.}
  \label{tab:FEVER-examples}
  \begin{tabular}{|c|p{3cm}|p{2.5cm}|p{5cm}|p{5cm}|}
    \hline
    \textbf{Ex \#} & \textbf{QUERY} & \textbf{Adversarial suffix} & \textbf{Original searched document} & \textbf{Retrieved (targeted) document} \\
    \hline
    ex1 & Jensen Ackles was nominated for several Daytime Emmy Awards. 
        & \#\#↓ contribute \#\#ons drowning brave 
        & Jensen Ross Ackles (born March 1, 1978) is an American actor and director. He is known for his roles in television, including Dean Winchester on “Supernatural.” He received Daytime Emmy nominations for his work on “Days of Our Lives.” 
        & The discography of the English rock band Embrace. The band’s highest charting single is “Nature’s Law,” which reached \#2 on the UK Singles Chart in 2006. \\
    \hline
    ex2 & Steve Buscemi the film “Trees Lounge”’s director. 
        & \#\#hul \#\#apes ] \#\#erine holes 
        & Steven Vincent Buscemi (born December 13, 1957) is an American actor and film director. Buscemi wrote and directed ``Trees Lounge'' (1996), which he also starred in. The film premiered at the Sundance Film Festival and won the Waldo Salt Screenwriting Award.
        & Kadapra is a village in Koipuram Panchayath of Pathanamthitta District, Kerala, India. It is known for the ancient Kadapra Shiva Temple and Iringhavala Kavu temple festivals. \\
    \hline
    ex3 & The \textit{Underworld} film series has grossed a total of \$539 million. 
        & staples reprinted appendix [unused42] alison 
        & Underworld is a series of action horror films directed by Len Wiseman, Patrick Tatopoulos, Måns Mårlind, and Björn Stein. The first film (2003) grossed \$95 million worldwide. The entire franchise has grossed over \$539 million globally. 
        & This article is about the particular significance of the year 1717 to Wales and its people. In 1717, the first Welsh-language newspaper was published, “Eurgrawn Cymraeg.” \\
    \hline
    ex4 & Aneurin Barnard’s birth year is 1987. 
        & chloride gp pretended pp dilapidated 
        & Aneurin Barnard (born 8 May 1987) is a Welsh stage and screen actor. He is best known for roles in “Dunkirk” (2017) and “The Goldfinch” (2019). He won the BAFTA Cymru Award for Best Actor in 2014. 
        & \textit{Tumi Kar?} (2008) is a Bengali film directed by Gobinda Chandra Haldar and produced by Agni Productions. It stars Dipankar Dey and Moushumi Chatterjee. \\
    \hline
    ex5 & Claudia Leitte is on “The Voice Brasil” as a coach and mentor. 
        & ruth evelyn kurdistan treaty granite 
        & Cl\'{a}udia Cristina Leite In\'{a}cio Pedreira (born 10 July 1980) is a Brazilian singer and songwriter. She served as a coach on ``The Voice Brasil'' from 2013 to 2016 and again in 2019.
        & Mukesh Dhirubhai Ambani (born 19 April 1957) is an Indian business magnate who is the chairman and managing director of Reliance Industries. He is currently the richest person in India. \\
    \hline
  \end{tabular}
\end{table}
\endgroup

\subsection{Examples on MS MARCO Dataset}\label{app:appendix:A4}
\begingroup
\setlength{\textfloatsep}{3pt}\setlength{\intextsep}{3pt}\setlength{\floatsep}{3pt}
\begin{table}[H]
  \centering
  \caption{Example DE prefix using user query and passages from the MS MARCO dataset, including suffix, original searched document, and retrieved (targeted) document.}
  \label{tab:MsMarco-examples}
  \begin{tabular}{|c|p{3cm}|p{2.5cm}|p{5cm}|p{5cm}|}
    \hline
    \textbf{Ex \#} & \textbf{QUERY} & \textbf{Adversarial suffix} & \textbf{Original searched document} & \textbf{Retrieved (targeted) document} \\
    \hline
    ex1 & Collagen, the most abundant protein in the body, plays a role in which function? 
        & rotary sprung awake prosecuted jessica 
        & Collagen is the major structural protein in connective tissue and the most abundant protein in the human body. It is responsible for skin elasticity, bone strength, tendon flexibility, and wound healing. 
        & Lisa Glasberg; Born: Lisa Glasberg June 6, 1956 (age 61); Other names: Lisa G; Alma mater: Hofstra University; Occupation: radio and TV personality at WFAN, Howard 101. \\
    \hline
    ex2 & Where is blood coming from when it enters the tricuspid valve and passes into the right ventricle? 
        & retailer staffed exchanging bread economical 
        & As deoxygenated blood flows into the right atrium, it passes through the tricuspid valve and into the right ventricle, which then pumps it into the pulmonary artery. 
        & I recommend this course. Thanks to: myalftraining.com online course (ALF Core Training), I have gained immense knowledge about Air Leakage Fundamentals. \\
    \hline
    ex3 & Which of these terms is defined as words that are spelled the same but have different meanings and pronunciations? 
        & honestly chen dowry belonging 
        & Homographs are words that are spelled the same but differ in meaning, derivation, or pronunciation (e.g., “lead” the metal vs. “lead” to guide). Homophones are words that sound the same but have different spellings (e.g., “red” vs. “read”). 
        & The Reasons You Should Be Sipping Apple Cider Vinegar: I drink apple cider vinegar just like you described, Kim, with lemon and honey mixed in for breakfast to aid digestion. \\
    \hline
    ex4 & What is the rhythmic wavelike muscular contraction of the esophagus that helps work the food down into the stomach called? 
        & waited chateau officials male revealing 
        & Digestive System – The esophagus is a muscular tube connecting from the throat to the stomach. The rhythmic wavelike contraction that propels food is called peristalsis. 
        & Local news and events from Whitefish Bay, WI Patch. Latest headlines: Winter Storm Snowfall Totals, Plus More Snow On The Way. \\
    \hline
    ex5 & What is the name of the condition characterized by disordered eating, lack of menstrual periods, and osteoporosis? 
        & 51st \#\#ße vacant kidnapped devastated 
        & But for some girls, not balancing the needs of their bodies and their sports can have major consequences. Some girls who exercise excessively and restrict calories may develop the “female athlete triad,” which includes eating disorders, amenorrhea, and osteoporosis. 
        & Koenigsegg. Bugatti’s Veyron. Lamborghini. Ferrari’s Pininfarina. The names alone are exotic. Here are the most expensive hypercars in the world. \\
    \hline
  \end{tabular}
\end{table}
\endgroup

\section{Influence of Suffix Length}
\label{app:suf_len}
\begin{figure}[t]
  \centering
  \begin{subfigure}[t]{0.48\linewidth}
    \centering
    \includegraphics[width=\linewidth]{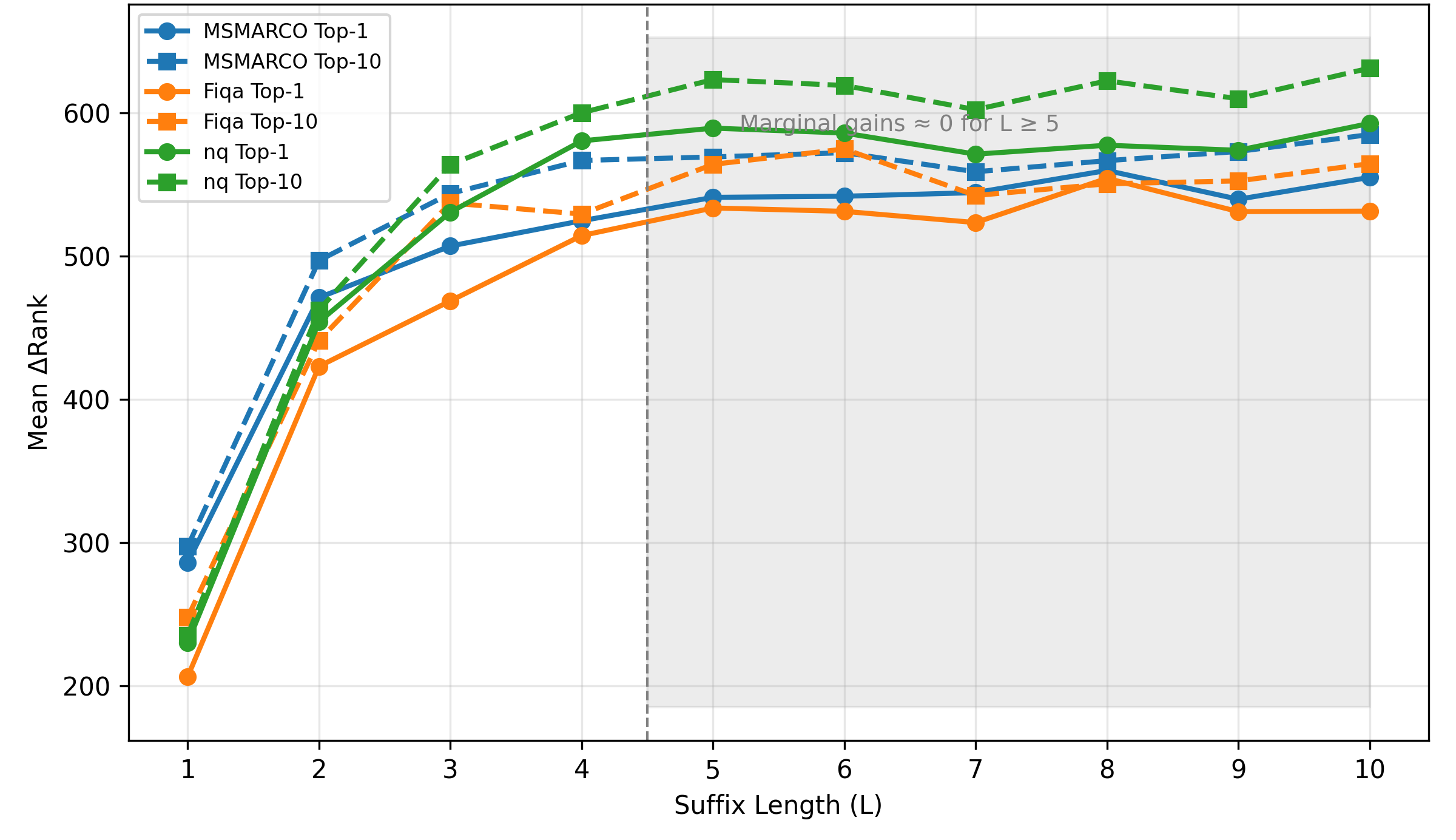}
    \caption{Mean $\Delta$Rank vs.\ suffix length (Top-1 / Top-10).}
    \label{fig:mean-deltarank}
  \end{subfigure}
  \hfill
  \begin{subfigure}[t]{0.48\linewidth}
    \centering
    \includegraphics[width=\linewidth]{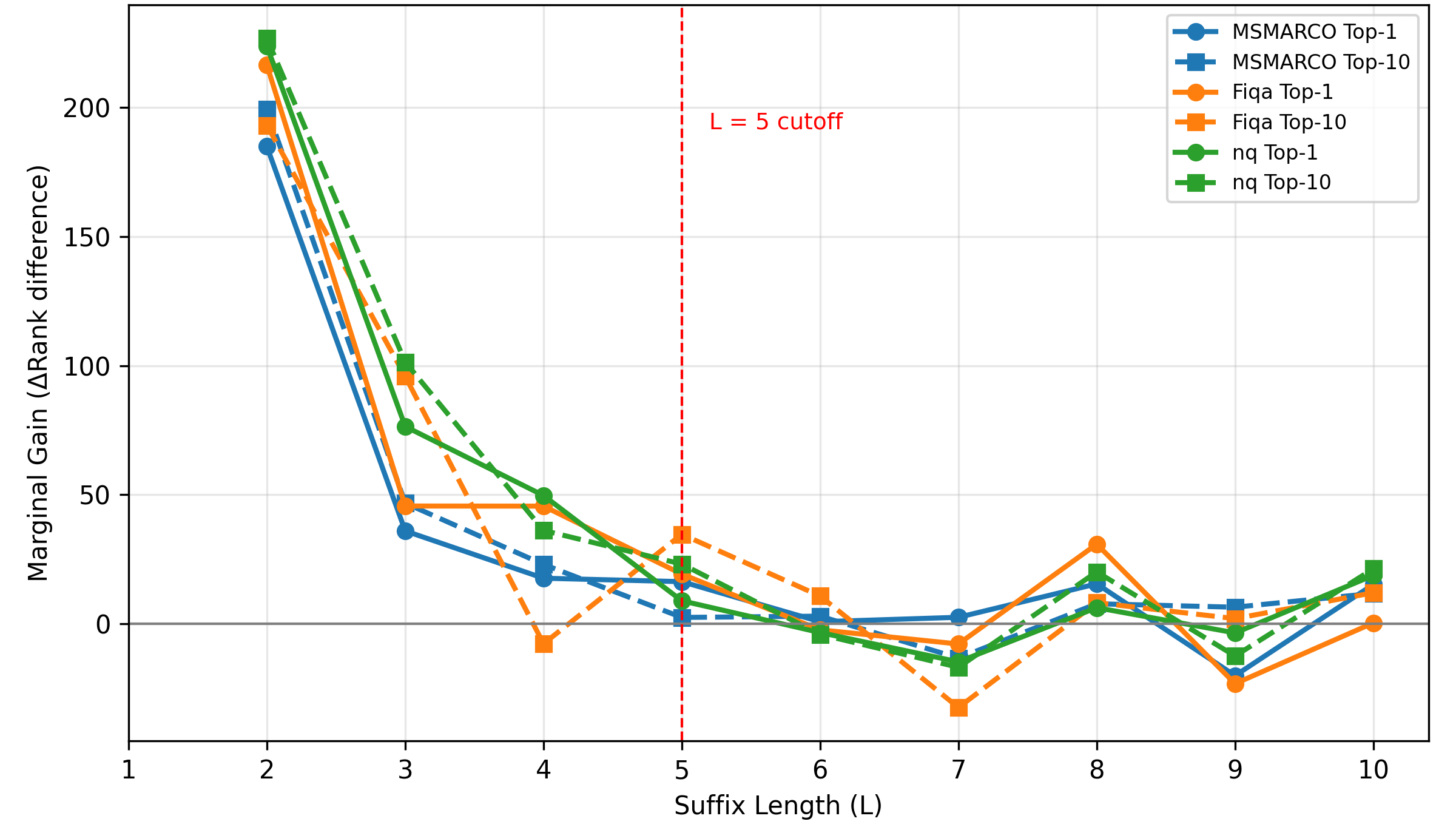}
    \caption{Marginal $\Delta$Rank gain per extra token (Top-1 / Top-10).}
    \label{fig:marginal-gain}
  \end{subfigure}
  \caption{(a) Mean $\Delta$Rank as suffix length increases across MS~MARCO, FiQA, and NQ. 
  (b) Marginal gain in $\Delta$Rank per additional suffix token.}
  \label{fig:combined-deltarank}
\end{figure}

To investigate the influence of suffix length on adversarial effectiveness in retrieval-based systems, we conduct experiments across three benchmark datasets—MS~MARCO, FiQA, and NQ~\cite{Kwiatkowski2019NQ}—under two ranking objectives: Top-1 and Top-10. Two key metrics are examined in this ablation study: the mean improvement in target document rank (\textit{Mean $\Delta$Rank}) and the incremental improvement obtained by increasing the suffix length by one token (\textit{Marginal Gain}). These metrics are visualized in Figure~\ref{fig:combined-deltarank}: panel~(a) plots Mean $\Delta$Rank and panel~(b) plots Marginal Gain. In this experiment, we assume that the target corpus is the one with rank $=800$, targeting the Top-1 or Top-10 result to make sure the maximum difference of the result stays the same and also ensure that there is sufficient variation to assess how the $\Delta\mathrm{Rank}$ efficient frontier diminishes.

Figure~\ref{fig:mean-deltarank} shows that for both Top-1 and Top-10 settings, $\Delta$Rank rises sharply from $L=1$ to $L=4$. Beyond $L=5$ (shaded region), further tokens yield negligible or negative gains, indicating a clear plateau in adversarial effectiveness.

To provide a more granular understanding of this phenomenon, we compute the marginal gain for each suffix length, as illustrated in Figure~\ref{fig:marginal-gain}. Marginal gain is defined as the difference in mean $\Delta\text{Rank}$ between suffix lengths $L$ and $L-1$, formally:
\begingroup
  \setlength\abovedisplayskip{4pt}
  \setlength\belowdisplayskip{4pt}
  \begin{equation}\label{eq:marginal-gain}
    \mathrm{MarginalGain}(L)
      = \mathrm{Mean}\,\Delta\mathrm{Rank}(L)
      - \mathrm{Mean}\,\Delta\mathrm{Rank}(L-1).
  \end{equation}
\endgroup
\label{sec:suffix-length}

This measure directly quantifies the effectiveness of adding one additional token at each step. All six curves exhibit a steep decline in marginal gain after $L=2$, reaching near zero by $L=5$. Although the absolute $\Delta\mathrm{Rank}$ values vary across datasets, the curves share the same shape and exhibit diminishing returns beyond five tokens, implying that, under a fixed iteration budget, attackers should craft concise perturbations rather than adding tokens indiscriminately.

\section{Aggregated Results on Loss Comparison}\label{app:tab:loss_aggregate_comparison}
To evaluate the relative effectiveness of Hinge Loss and Cosine Loss in adversarial tail-patch attacks, we conducted experiments on four benchmarks. For every query, we selected as the adversarial target the relevant document that was ranked exactly 100th under the original cosine-similarity baseline. The goal of the attack was to promote this target document to the top of the ranking specifically, from rank 100 to rank 1 by appending a five-token adversarial suffix to each query. Differential Evolution was used to optimize the suffix, minimizing either a cosine-loss objective or a hinge-based ranking objective. Let the embedding of the adversarially modified query be denoted as $\bm{u} \in \mathbb{R}^d$, and the embedding of the target document as $\bm{v} \in \mathbb{R}^d$. 
Cosine Loss is defined as the negative direction of this similarity:
\begingroup
  \setlength\abovedisplayskip{4pt}
  \setlength\belowdisplayskip{4pt}
  \begin{equation}\label{eq:cosine-loss}
    \mathcal{L}_{\text{cos}}(\bm{u}, \bm{v})
      = -\cos(\bm{u}, \bm{v})
      = -\frac{\bm{u}^\top \bm{v}}
               {\lVert \bm{u} \rVert \,\lVert \bm{v} \rVert}.
  \end{equation}
\endgroup
While hinge loss is defined above.
In optimization, minimizing $\mathcal{L}_{\text{cos}}$ is equivalent to directly maximizing the alignment between the query and the target document in embedding space.
However, as shown in As shown in Table~\ref{tab:loss_aggregate_comparison}, the cosine objective yields worse ranking performance than the hinge loss while incurring greater—and undesirable—semantic drift, and still does not reliably place the target document  to the top rank.
\begin{table}[ht]
\centering
\caption{Aggregated Results for Hinge Loss versus Cosine Loss (baseline target rank = 100). All values are averaged over 100 queries per dataset.}
\label{tab:loss_aggregate_comparison}
\resizebox{\textwidth}{!}{%
\begin{tabular}{l
S[table-format=1.4] S[table-format=1.4]
S[table-format=2.2] S[table-format=2.2]
S[table-format=4.0] S[table-format=4.0]
S[table-format=1.4] S[table-format=1.4]}
\toprule
Dataset &
\multicolumn{1}{c}{Cosine $\Delta \cos$} &
\multicolumn{1}{c}{Hinge $\Delta \cos$} &
\multicolumn{1}{c}{Cosine $\Delta \text{rank}$} &
\multicolumn{1}{c}{Hinge $\Delta \text{rank}$} &
\multicolumn{1}{c}{Cosine Iters} &
\multicolumn{1}{c}{Hinge Iters} &
\multicolumn{1}{c}{Cosine Succ.\ Rate} &
\multicolumn{1}{c}{Hinge Succ.\ Rate} \\
\midrule
FEVER      & 0.0686 & -0.0273 & 59.04 & 76.26 & 4425 & 3615 & 0.0100 & 0.2600 \\
FiQA-2018  & 0.0422 & -0.0092 & 63.92 & 86.99 & 4274 & 3743 & 0.0200 & 0.2300 \\
MS MARCO   & 0.0634 &  0.0208 & 77.63 & 88.25 & 4346 & 3499 & 0.0700 & 0.2900 \\
SciFact    & 0.0577 & -0.0203 & 60.67 & 80.94 & 4624 & 3466 & 0.0200 & 0.3100 \\
\bottomrule
\end{tabular}%
}
\end{table}

\section{Distribution plots}
\label{appendix:distributions}
\begin{table}[H]
    \centering
    \caption{Detector means for the original query~(Q) and the
             attacked query with injection
             (Q$\parallel$S).}
    \label{tab:de-det-stats}\small
    \begin{tabular}{@{\hspace{2pt}}lcccc@{}}
        \toprule
        \multirow{2}{*}{\textbf{Dataset}}
        & \multicolumn{2}{c}{\textbf{PPL}} & \multicolumn{2}{c}{\textbf{CLS prob.}}\\
        \cmidrule(lr){2-3}\cmidrule(l){4-5}
        & Q & Q$\parallel$S & Q & Q$\parallel$S\\
        \midrule
        FEVER    & 1.451 & 1.491 & 0.407 & 0.402 \\
        FIQA     & 2.053 & 1.544 & 0.393 & 0.400 \\
        MS MARCO & 42.57 & 1.620 & 0.399 & 0.399 \\
        SciFact  & 1.380 & 1.419 & 0.403 & 0.401 \\
        \bottomrule
    \end{tabular}
\end{table}

\begin{figure}[h]
    \centering
    \includegraphics[width=\textwidth]{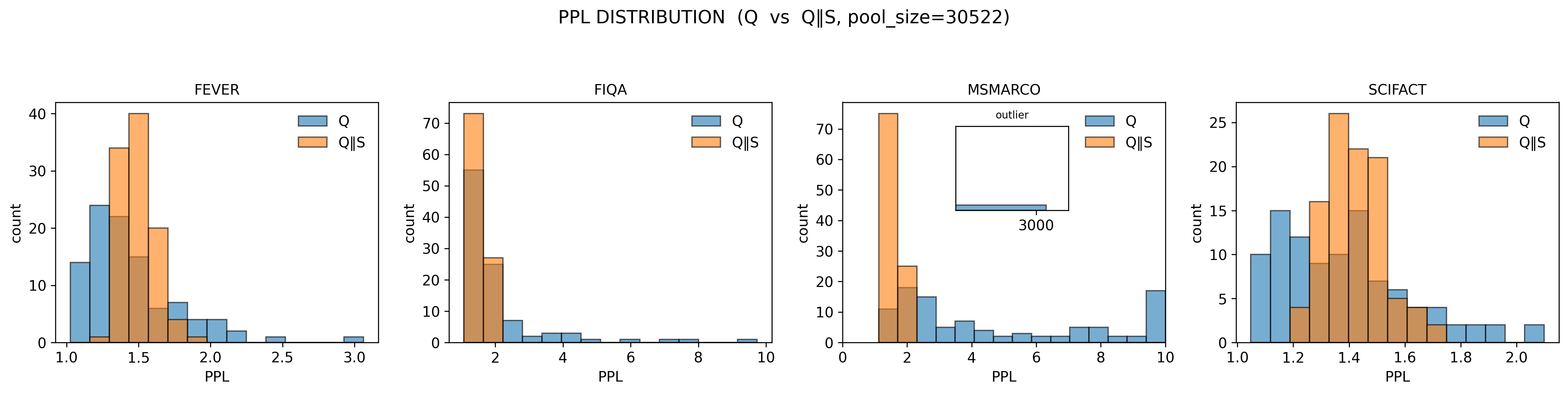}
    \caption{\textbf{PPL distributions}.
      Blue~=~original queries~(Q); orange~=~attacked
      queries~(Q$\parallel$S, pool size ${\approx}30$\,k).  
      The inset on the \textsc{MS-MARCO} panel highlights the single
      extreme outlier (${\approx}3{,}000$).}
    \label{fig:ppl-dist}
\end{figure}

\begin{figure}[h]
    \centering
    \includegraphics[width=\textwidth]{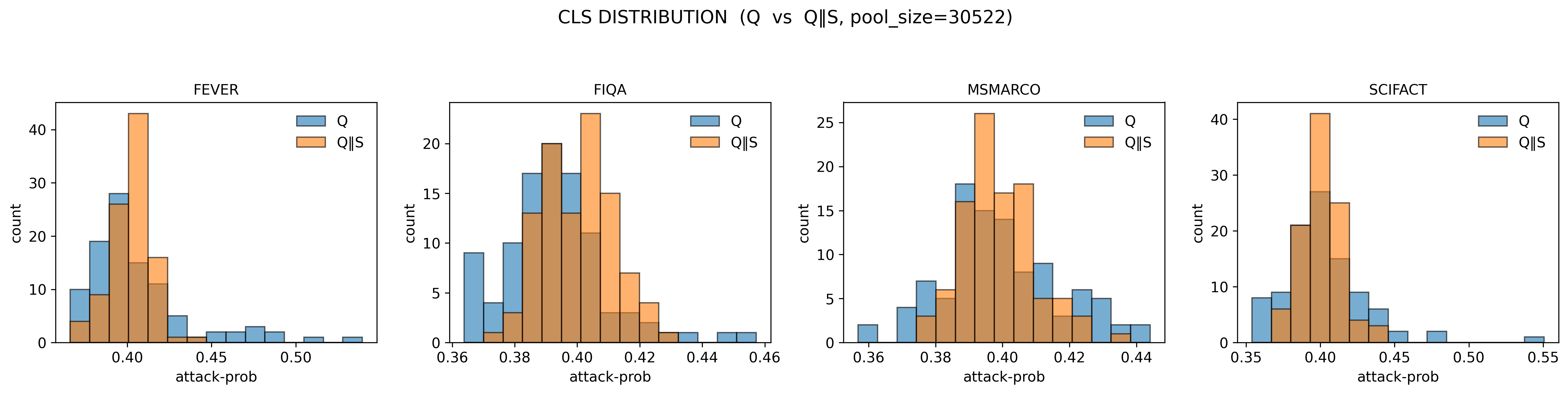}
    \caption{\textbf{CLS attack-probability distributions} for the same
      100 query pairs as Fig.~\ref{fig:ppl-dist}.}
    \label{fig:cls-dist}
\end{figure}

\section{Pooling Time on MLM Strategy}
\label{app:pooling-time}

Table~\ref{tab:time_metrics} reports pool construction (“Build”) and query optimization (“Query”) times. Total attack time is their sum. The results demonstrate the efficiency of constructing a readability‑aware candidate pool to improve suffix quality.
\begin{table}[!htbp]
\centering
\caption{Pool construction (“Build”) and query optimization (“Query”) times (seconds). Total attack time is the sum of the two.}
\small
\begin{tabular}{c|cc|cc|cc}
\hline
\bfseries Pool Size 
  & \multicolumn{2}{c|}{\bfseries FEVER (s)}
  & \multicolumn{2}{c|}{\bfseries FiQA (s)}
  & \multicolumn{2}{c}{\bfseries SciFact (s)} \\
  & Build & Query 
  & Build & Query 
  & Build & Query \\
\hline
500     & 0.0117 & 44.4184 & 0.0121 & 51.0981 & 0.0106 & 50.4705 \\
1\,000  & 0.0131 & 47.2129 & 0.0128 & 43.0398 & 0.0106 & 48.6922 \\
2\,000  & 0.0132 & 44.3453 & 0.0132 & 46.2533 & 0.0107 & 44.8168 \\
5\,000  & 0.0131 & 45.2823 & 0.0114 & 41.4641 & 0.0117 & 45.1493 \\
10\,000 & 0.0153 & 46.0093 & 0.0113 & 40.9918 & 0.0159 & 42.6680 \\
20\,000 & 0.0155 & 39.5988 & 0.0122 & 46.6468 & 0.0144 & 45.9617 \\
30\,522 & 0.0004 & 45.5934 & 0.0004 & 41.4091 & 0.0004 & 43.4744 \\
\hline
\end{tabular}
\label{tab:time_metrics}
\end{table}

\section{Additional Readability Analysis}
\label{app:readability}

Table~\ref{tab:nll} shows the average MLM negative log‐likelihood (NLL) of generated suffixes under each pool size, where NLL clearly decreases as the pool shrinks. MLM NLL is often treated as a readability proxy. Table~\ref{tab:welch} reports Welch’s $t$‐test comparing $\mathrm{pool\_size}=5{,}000$ vs.\ full ($30{,}522$), confirming the reduction is significant across all datasets. Together, these results show the lightweight pooling strategy improves suffix fluency with negligible attack‑success impact.

\begin{table}[t]
\centering
\caption{Average MLM NLL of generated suffixes under different pool sizes.}
\small
\begin{tabular}{c|ccc}
\hline
\bfseries Pool size & \bfseries Fever NLL & \bfseries FiQA NLL & \bfseries SciFact NLL \\
\hline
500     & $7.13 \pm 1.26$ & $6.65 \pm 1.17$ & $7.19 \pm 1.03$ \\
1\,000  & $7.19 \pm 1.03$ & $7.06 \pm 1.13$ & $7.42 \pm 0.91$ \\
2\,000  & $7.48 \pm 1.03$ & $7.48 \pm 1.03$ & $7.65 \pm 0.94$ \\
5\,000  & $7.78 \pm 0.86$ & $7.90 \pm 1.00$ & $7.78 \pm 0.77$ \\
10\,000 & $8.17 \pm 0.97$ & $8.06 \pm 0.79$ & $8.31 \pm 0.82$ \\
20\,000 & $8.55 \pm 0.94$ & $8.57 \pm 1.00$ & $8.66 \pm 0.86$ \\
30\,522 & $8.78 \pm 0.85$ & $8.77 \pm 0.86$ & $8.84 \pm 0.81$ \\
\hline
\end{tabular}
\label{tab:nll}
\end{table}

\begin{table}[t]
\centering
\caption{Welch’s $t$-test comparing MLM NLL for pool\_size $=5{,}000$ vs.\ full pool ($30{,}522$).}
\small
\begin{tabular}{l|rr}
\hline
\bfseries Dataset & \bfseries $t$-value & \bfseries $p$-value \\
\hline
Fever   & $-8.33$ & $1.34\times10^{-14}$ \\
FiQA    & $-6.54$ & $5.33\times10^{-10}$ \\
SciFact & $-9.40$ & $1.39\times10^{-17}$ \\
\hline
\end{tabular}
\label{tab:welch}
\end{table}



\begin{thebibliography}{99}

\bibitem{bajaj2018msmarcohumangenerated}
Payal Bajaj, Daniel Campos, Nick Craswell, Li Deng, Jianfeng Gao, Xiaodong Liu,
Rangan Majumder, Andrew McNamara, Bhaskar Mitra, Tri Nguyen, Mir Rosenberg,
Xia Song, Alina Stoica, Saurabh Tiwary, and Tong Wang.
\newblock MS MARCO: A Human Generated Machine Reading Comprehension Dataset.
\newblock \emph{arXiv preprint arXiv:1611.09268}, 2018.
\newblock \url{https://arxiv.org/abs/1611.09268}.

\bibitem{Maia2018Fiqa}
Macedo Maia, Siegfried Handschuh, Andre Freitas, Brian Davis, Ross McDermott,
Manel Zarrouk, and Alexandra Balahur.
\newblock WWW’18 open challenge: Financial opinion mining and question answering.
\newblock In \emph{Companion of the The Web Conference 2018 (WWW ’18 Companion)},
pages 1941--1942, Lyon, France, April 2018.
\newblock International World Wide Web Conferences Steering Committee.


\bibitem{SciFact2020}
Daniel Wadden, Rishi Bommasani, Russell Kaplan, Swabha Swayamdipta, and
Nanyun Peng.
\newblock Fact or Fiction: Verifying Scientific Claims.
\newblock In \emph{Proceedings of EMNLP}, pages —, 2020.

\bibitem{thorne2018fever}
James Thorne, Andreas Vlachos, Christos Christodoulopoulos, and Arpit Mittal.
\newblock FEVER: A large-scale dataset for fact extraction and VERification.
\newblock In \emph{Proc. of NAACL-HLT}, pages 809--819, New Orleans, LA, USA,
2018.
\newblock \url{https://aclanthology.org/N18-1074/}.
\bibitem{rajpurkar2016squad}
Pranav Rajpurkar, Jian Zhang, Konstantin Lopyrev, and Percy Liang.
\newblock SQuAD: 100{,}000+ questions for machine comprehension of text.
\newblock In \emph{Proc. of EMNLP}, pages 2383--2392, Austin, TX, USA, 2016.
\newblock \url{https://aclanthology.org/D16-1264/}.
\bibitem{Thakur2021Beir}
Nandan Thakur, Joel Mackenzie, Hasan Sajjad, Ziyu Yao, and Preslav Nakov.
\newblock BEIR: A Heterogeneous Benchmark for Zero-shot Evaluation of
Information Retrieval Models.
\newblock \emph{arXiv preprint arXiv:2104.08663}, 2021.
\newblock \url{https://arxiv.org/abs/2104.08663}.
\bibitem{Kwiatkowski2019NQ}
Tom Kwiatkowski, Jennimaria Palomaki, Olivia Redfield, Michael Collins, Ankur Parikh, Chris Alberti, Danielle Epstein, Illia Polosukhin, Jacob Devlin, Kenton Lee, Kristina Toutanova, Llion Jones, Matthew Kelcey, Ming-Wei Chang, Andrew M. Dai, Jakob Uszkoreit, Quoc Le, and Slav Petrov.  
\newblock Natural Questions: A Benchmark for Question Answering Research.  
\newblock \emph{Transactions of the Association for Computational Linguistics}, 7:452--466, 2019.  
\newblock \url{https://aclanthology.org/Q19-1026/}.

\bibitem{DE1997}
Rainer Storn and Kenneth Price.
\newblock Differential evolution – a simple and efficient heuristic for global
optimization over continuous spaces.
\newblock \emph{Journal of Global Optimization}, 11(4):341–359, 1997.

\bibitem{DEReview2021}
Mohamed Alsayed Elaziz, Ahmed Ali Ismail, Ashraf M. Salem, and Ahmed E. Hassanien.
\newblock Differential Evolution: A Recent Review Based on State-of-the-Art
Works.
\newblock \emph{Alexandria Engineering Journal}, 60(4):3595–3615, 2021.

\bibitem{li2024enhancingllmfactualaccuracy}
Jiarui Li, Ye Yuan, and Zehua Zhang.
\newblock Enhancing LLM Factual Accuracy with RAG to Counter Hallucinations:
A Case Study on Domain-Specific Queries in Private Knowledge-Bases.
\newblock \emph{arXiv preprint arXiv:2403.10446}, 2024.
\newblock \url{https://arxiv.org/abs/2403.10446}.

\bibitem{li2024}
Xuying Li, Zhuo Li, Yuji Kosuga, Yasuhiro Yoshida, and Victor Bian.
\newblock Targeting the Core: A Simple and Effective Method to Attack
RAG-based Agents via Direct LLM Manipulation.
\newblock \emph{arXiv preprint arXiv:2412.04415}, 2024.
\newblock \url{https://arxiv.org/abs/2412.04415}.

\bibitem{xue2024}
Jiaqi Xue, Mengxin Zheng, Yebowen Hu, Fei Liu, Xun Chen, and Qian Lou.
\newblock BadRAG: Identifying Vulnerabilities in Retrieval Augmented Generation
of Large Language Models.
\newblock \emph{arXiv preprint arXiv:2406.00083}, 2024.
\newblock \url{https://arxiv.org/abs/2406.00083}.
\bibitem{alon2023detecting}
Gabriel Alon and Michael Kamfonas.  
\newblock Detecting Language Model Attacks with Perplexity.  
\newblock \emph{arXiv preprint arXiv:2308.14132}, 2023.  
\newblock \url{https://arxiv.org/abs/2308.14132}.

\bibitem{kim2024aps}
Jinhwa Kim, Ali Derakhshan, and Ian Harris.  
\newblock Robust Safety Classifier Against Jailbreaking Attacks: Adversarial Prompt Shield.  
\newblock In \emph{Proceedings of the 8th Workshop on Online Abuse and Harms (WOAH 2024)}, pages 159--170, Mexico City, Mexico, June 2024. Association for Computational Linguistics.  
\newblock \url{https://aclanthology.org/2024.woah-1.12}.
\bibitem{sui2025ctrlrag}
Runqi Sui.
\newblock CtrlRAG: Black-box Adversarial Attacks Based on Masked Language
Models in Retrieval-Augmented Language Generation.
\newblock \emph{arXiv preprint arXiv:2503.06950}, 2025.
\newblock \url{https://arxiv.org/abs/2503.06950}.
\bibitem{wu2022prada}
Chen Wu and Ruqing Zhang.
\newblock PRADA: Practical Black-box Adversarial Attacks against Neural Ranking Models.
\newblock \emph{ACM Transactions on Information Systems}, 41 (Dec. 16, 2022).
\newblock \url{https://doi.org/10.1145/3576923}.
\bibitem{Su_2019}
Jiawei Su, Danilo Vasconcellos Vargas, and Kouichi Sakurai.
\newblock One Pixel Attack for Fooling Deep Neural Networks.
\newblock \emph{IEEE Transactions on Evolutionary Computation}, 23(5):828–841,
2019.
\newblock \url{https://doi.org/10.1109/TEVC.2019.2890858}.

\bibitem{devlin2018bert}
Jacob Devlin, Ming-Wei Chang, Kenton Lee, and Kristina Toutanova.
\newblock BERT: Pre-training of Deep Bidirectional Transformers for Language
Understanding.
\newblock \emph{arXiv preprint arXiv:1810.04805}, 2018.
\newblock \url{http://arxiv.org/abs/1810.04805}.

\bibitem{perez2022}
Fábio Perez and Ian Ribeiro.
\newblock Ignore Previous Prompt: Attack Techniques for Language Models.
\newblock \emph{arXiv preprint arXiv:2211.09527}, 2022.
\newblock \url{https://arxiv.org/abs/2211.09527}.

\bibitem{liu2024promptinjection}
Xiaogeng Liu, Zhiyuan Yu, Yizhe Zhang, Ning Zhang, and Chaowei Xiao.
\newblock Automatic and Universal Prompt Injection Attacks Against Large
Language Models.
\newblock \emph{arXiv preprint arXiv:2403.04957}, 2024.
\newblock \url{https://arxiv.org/abs/2403.04957}.

\bibitem{zhang2024promptinjection}
Chong Zhang, Mingyu Jin, Qinkai Yu, Chengzhi Liu, Haochen Xue, and Xiaobo Jin.
\newblock Goal-guided Generative Prompt Injection Attack on Large Language
Models.
\newblock \emph{arXiv preprint arXiv:2404.07234}, 2024.
\newblock \url{https://arxiv.org/abs/2404.07234}.

\bibitem{chao2024jailbreaking}
Patrick Chao, Alexander Robey, Edgar Dobriban, Hamed Hassani, George J. Pappas,
and Eric Wong.
\newblock Jailbreaking Black Box Large Language Models in Twenty Queries.
\newblock \emph{arXiv preprint arXiv:2310.08419}, 2024.
\newblock \url{https://arxiv.org/abs/2310.08419}.

\bibitem{Hu2024GGPP}
Zhibo Hu, Chen Wang, Yanfeng Shu, Hye-young Paik, and Liming Zhu.
\newblock Prompt Perturbation in Retrieval-Augmented Generation based Large
Language Models.
\newblock \emph{arXiv preprint arXiv:2402.07179}, 2024.
\newblock \url{https://arxiv.org/abs/2402.07179}.

\bibitem{zhou2025tempest}
Andy Zhou and Ron Arel.
\newblock TEMPEST: Multi-Turn Jailbreaking of Large Language Models with Tree
Search.
\newblock In \emph{Proceedings of the ICLR 2025 Workshop on Building Trust in
Language Models and Applications}, 2025.
\newblock \url{https://openreview.net/forum?id=rDC2UVdB0t}.


\bibitem{lewis2021nlp}
Patrick Lewis, Ethan Perez, Aleksandra Piktus, Fabio Petroni, Vladimir
Karpukhin, Naman Goyal, Heinrich Küttler, Mike Lewis, Wen-tau Yih, Tim
Rocktäschel, Sebastian Riedel, and Douwe Kiela.
\newblock Retrieval-Augmented Generation for Knowledge-Intensive NLP Tasks.
\newblock \emph{arXiv preprint arXiv:2005.11401}, 2021.
\newblock \url{https://arxiv.org/abs/2005.11401}.

\newblock \url{https://arxiv.org/abs/2011.03901}.
\bibitem{Solaiman2019}
Irene Solaiman, Miles Brundage, Jack Clark, Amanda Askell, Ariel Herbert‐Voss, Jeff Wu, Alec Radford, Gretchen Krueger, Jong Wook Kim, Sarah Kreps, and others.
\newblock Release strategies and the social impacts of language models.
\newblock In \emph{arXiv preprint arXiv:1908.09203}, pages —, 2019.

\bibitem{guo2025evoprompt}
Qingyan Guo, Rui Wang, Junliang Guo, Bei Li, Kaitao Song, Xu Tan, Guoqing
Liu, Jiang Bian, and Yujiu Yang.
\newblock EvoPrompt: Connecting LLMs with Evolutionary Algorithms Yields
Powerful Prompt Optimizers.
\newblock \emph{arXiv preprint arXiv:2309.08532}, 2025.
\newblock \url{https://arxiv.org/abs/2309.08532}.

\bibitem{lin2020blackbox}
Junyu Lin, Lei Xu, Yingqi Liu, and Xiangyu Zhang.
\newblock Black-box Adversarial Sample Generation Based on Differential
Evolution.
\newblock \emph{arXiv preprint arXiv:2007.15310}, 2020.
\newblock \url{https://arxiv.org/abs/2007.15310}.

\bibitem{diao2023blackbox}
Shizhe Diao, Zhichao Huang, Ruijia Xu, Xuechun Li, Yong Lin, Xiao Zhou, and
Tong Zhang.
\newblock Black-Box Prompt Learning for Pre-trained Language Models.
\newblock \emph{Transactions on Machine Learning Research}, to appear, 2023.
\newblock \url{https://openreview.net/forum?id=IvsGP7xRvm}.


\bibitem{muennighoff2024grit}
Niklas Muennighoff, Hongjin Su, Liang Wang, Nan Yang, Furu Wei, Tao Yu,
Amanpreet Singh, and Douwe Kiela.
\newblock Generative Representational Instruction Tuning.
\newblock \emph{arXiv preprint arXiv:2402.09906}, 2024.
\newblock \url{https://arxiv.org/abs/2402.09906}.

\bibitem{merrick2024arcticembed}
Luke Merrick, Danmei Xu, Gaurav Nuti, and Daniel Campos.
\newblock Arctic-Embed: Scalable, Efficient, and Accurate Text Embedding
Models.
\newblock \emph{arXiv preprint arXiv:2405.05374}, 2024.
\newblock \url{https://arxiv.org/abs/2405.05374}.

\bibitem{hu2023token}
Zhengmian Hu, Yichao Lu, Xuejun Liao, Jianfeng Gao, and Maxine Eskenazi.
\newblock Token-Level Adversarial Prompt Detection Based on Perplexity
Measures and Contextual Information.
\newblock \emph{arXiv preprint arXiv:2311.11509}, 2023.
\newblock \url{https://arxiv.org/abs/2311.11509}.

\bibitem{jha2025harnessing}
Rishi Jha, Collin Zhang, Vitaly Shmatikov, and John X. Morris.
\newblock Harnessing the Universal Geometry of Embeddings.
\newblock \emph{arXiv preprint arXiv:2505.12540}, 2025.
\newblock \url{https://arxiv.org/abs/2505.12540}.
\bibitem{pen9rum2025RagAttackDeRag}
pen9rum,
\emph{Rag\_attack\_DeRag: Differential Evolution Attacks on Retrieval-Augmented Generation},
GitHub repository,
\url{https://github.com/pen9rum/Rag_attack_DeRag},
accessed May 25, 2025.
\bibitem{robertson2009probabilistic}
Stephen Robertson and Hugo Zaragoza.
\newblock The Probabilistic Relevance Framework: BM25 and Beyond.
\newblock \emph{Foundations and Trends in Information Retrieval}, 3(1):333--389, 2009.
\newblock \url{https://doi.org/10.1561/1500000019}.

\bibitem{karpukhin2020dense}
Vladimir Karpukhin, Barlas Oguz, Sewon Min, Patrick Lewis, Ledell Wu, Sergey Edunov, Danqi Chen, and Wen-tau Yih.
\newblock Dense Passage Retrieval for Open-Domain Question Answering.
\newblock In \emph{Proceedings of the 2020 Conference on Empirical Methods in Natural Language Processing (EMNLP)}, pages 6769--6781, Online, 2020.
\newblock \url{https://aclanthology.org/2020.emnlp-main.550/}.

\end{thebibliography}
\end{document}